\documentclass{article}

\usepackage{PRIMEarxiv}

\usepackage[utf8]{inputenc} % allow utf-8 input
\usepackage[T1]{fontenc}    % use 8-bit T1 fonts
\usepackage{hyperref}       % hyperlinks
\usepackage{url}            % simple URL typesetting
\usepackage{booktabs}       % professional-quality tables
\usepackage{amsfonts}       % blackboard math symbols
\usepackage{nicefrac}       % compact symbols for 1/2, etc.
\usepackage{microtype}      % microtypography
\usepackage{lipsum}
\usepackage{fancyhdr}       % header
\usepackage{graphicx}       % graphics
\graphicspath{{media/}}     % organize your images and other figures under media/ folder
\usepackage[natbibapa]{apacite}
\usepackage{graphicx}
\graphicspath{ {./} }
\usepackage{multirow}
\usepackage[section]{placeins}
\usepackage{subfig}
\usepackage[normalem]{ulem}
\usepackage{booktabs}
\usepackage{array}
\usepackage{longtable}

\let\turc\c
\usepackage{etoolbox}
\robustify\turc

\title{Transparency challenges in policy evaluation with causal machine learning --- improving usability and accountability
}

\author{
  Patrick Rehill \\
  Centre for Social Research and Methods \\
  Australian National University \\
  Canberra\\
  \texttt{patrick.rehill@anu.edu.au} \\
   \And
  Nicholas Biddle \\
  Centre for Social Research and Methods \\
  Australian National University \\
  Canberra\\
  \texttt{nicholas.biddle@anu.edu.au} \\
}

\begin{document}
\maketitle

\begin{abstract}
Causal machine learning tools are beginning to see use in real-world policy evaluation tasks to flexibly estimate treatment effects. One issue with these methods is that the machine learning models used are generally black boxes, i.e., there is no globally interpretable way to understand how a model makes estimates. This is a clear problem in policy evaluation applications, particularly in government, because it is difficult to understand whether such models are functioning in ways that are fair, based on the correct interpretation of evidence and transparent enough to allow for accountability if things go wrong. However, there has been little discussion of transparency problems in the causal machine learning literature and how these might be overcome. This paper explores why transparency issues are a problem for causal machine learning in public policy evaluation applications and considers ways these problems might be addressed through explainable AI tools and by simplifying models in line with interpretable AI principles. It then applies these ideas to a case-study using a causal forest model to estimate conditional average treatment effects for a hypothetical change in the school leaving age in Australia. It shows that existing tools for understanding black-box predictive models are poorly suited to causal machine learning and that simplifying the model to make it interpretable leads to an unacceptable increase in error (in this application). It concludes that new tools are needed to properly understand causal machine learning models and the algorithms that fit them.
\end{abstract}

\section*{Policy significance statement}
Causal machine learning is beginning to be used in analysis that informs public policy. Particular techniques which estimate individual or group-level effects of interventions are the focus of this paper. The paper identifies two problems with applying causal machine learning to policy analysis --- usability and accountability issues, both of which require greater transparency in models. It argues that some existing tools can help to address these challenges but that users need to be aware of transparency issues and address them to the extent they can using the techniques in this paper. To the extent they cannot address issues, users need to decide whether more powerful estimation is really worth less transparency.

% keywords can be removed

\section{Introduction}
Causal machine learning is currently experiencing a surge of interest as a tool for policy evaluation \citep{CaglayanAkay2022Bibliometric,Lechner2023Causal}. With this enthusiasm and maturing of methods, we are likely to see more research using these methods that affect policy decisions. The promise of causal machine learning is that researchers performing causal estimation will be able to take advantage of machine learning models that have previously only been available to predictive modellers \citep{Imbens2021Breiman's,Daoud2020Statistical,Baiardi2021Value,Athey2017State}. Where traditional (supervised) predictive machine learning aims to estimate outcomes, causal machine learning aims to estimate treatment effects (the difference between an observed outcome for prediction and one which is fundamentally unobservable for causal modelling as the treatment effect will always be a function of an unobserved potential outcome \citep{Imbens2015Causal}). This generally means either plugging standard machine learning models into a special causal estimator or modifying machine learning methods to give causal estimates with good statistical properties (particularly asymptotic unbiasedness and consistency). This allows researchers to capture complex functional forms in high-dimensional data which relate cause to effect \citep{Chernozhukov2018Double/debiased,Knaus2021Double} and allows for a data-driven approach to estimate heterogeneous treatment effects that does not require explicitly including interactions with treatment \citep{Wager2018Estimation,Athey2019Generalized}. A good non-technical introduction to this literature can be found in \citet{Lechner2023Causal}.\footnote{We include a table of definitions in Appendix C as this paper uses many terms that will be unfamiliar to those without background knowledge in causal machine learning.}

There seems to be substantial benefits to using causal machine learning when the appropriate methods are applied correctly to the right research project. However, the fact that these methods generally use black-box models makes them very different from traditional causal estimation models. A model being 'black-box' means that it is not possible to get a general explanation of how a model arrived at an estimate \citep{Rudin2019Stop}. For example, in a linear regression, we can easily see how each coefficient multiplied by the data then summed gives a prediction, but in a model like a random forest, we need to understand the average result of potentially thousands of individual decision trees which is practically impossible. A black-box model then is one where we lack a reasonable general explanation of the functioning of the model, instead all that we can find are local explanations for how a particular prediction was made (later we will call this explainable AI (XAI) \citep{tang_explainable_2019}) or abandon the method and simplify to a ‘white-box’ model like a single decision tree (what we will later call interpretable AI (IAI) \citep{Rudin2019Stop}). This lack of a general explanation presents challenges when using causal machine learning methods to inform decision-making.

The focus of this paper will be on transparency in the case of heterogeneous treatment effect analysis in policy evaluation to inform public policy decisions. By transparency we mean an ability to get useful information about the workings of a black-box model. Specifically we focus on the causal forest method \citep{Athey2019Generalized,Wager2018Estimation}. We identify two kinds of transparency that are important, but which need to be thought of separately. These are termed accountability and usability. This classification of types of transparency is orthogonal to the means we might use to achieve transparency such as through XAI and IAI methods and in the latter half of this paper we discuss both types of methods as means for achieving both goals.

Accountability is transparency for those who will be subject to policy. Their interest in understanding the analysis used in policy-making is close to the classic case for transparency in predictive machine learning (see \citealt{IreniSaban2022Ethical} for an introduction to the literature on ethical issues around predictive machine learning and government) where a party subject to the decisions made by a model might be owed an explanation for the decisions made and the ability to identify and criticise injustices such as the right codified into the European Union’s GDPR \citep{Kim2022Why}. Specifically, transparency with an accountability goal is often concerned with addressing similar problems to machine learning fairness, though through the means of transparency rather than the often blunter means of fairness rules \citep{rai2020explainable}. This means that accountability concerns are often particularly focused on the use of sensitive variables like gender or race in models. However, this analogy to the predictive case is complicated somewhat by the role of the human decision-maker who is generally interpreting the results of a causal machine learning analysis and making decisions based on it \citep{rehill_fairness_2023}. Causal machine learning models would rarely make decisions directly as they might in predictive applications, but instead inform a longer policy-making process. It is necessary then to understand the output of a model, but it is also necessary to understand the human decision-making process that was informed by the output and which led to a policy outcome.

Usability is transparency that helps the analyst and decision-maker to understand the data generating process (DGP) and therefore obtain better insights into the causal processes at play. It can also help to diagnose problems in modelling for example, finding variables that are 'bad controls' \citep{hunermund_causal_2021} that should not be in the dataset. As with accountability, the primary difference between causal and predictive applications from a justice perspective is not the actual differences in estimation processes, but rather it is the way that causal models are generally there to inform human decision-makers while predictive ones generally exercise more direct power \citep{rehill_fairness_2023}. Usability is precisely the way in which models do this informing, taking a model of hundreds of thousands of parameters in the case of a typical causal forest and presenting the patterns in those parameters in a way that can tell the user about the underlying causal effects. Because usability is so directly tied to the human role, there is less of a parallel here to the existing transparency literature than there is in accountability but we will explore how existing transparency tools can still be useful for improving usability.

This paper is an effort to lay out the problems posed by applying black-box models to causal inference where methods have generally been interpretable in the past. There is little existing literature in this area (we are not familiar with any aside from \citet{GurAli2022Targeting}), however the critical literature around predictive learning provides a blueprint for understanding these concerns and trying to solve them. Section \ref{sec:introducing-causal-ml} provides a background on causal machine learning. Section \ref{sec:rationale} explains why these methods might be useful for policy-making. Section \ref{sec:transparency-needs} looks specifically at transparency in causal machine learning and the role of accountability and usability. Section \ref{sec:case-study} introduces the case study that will motivate the rest of the paper, a study of returns on education in Australia using the Household Income and Labour Dynamics in Australia Survey (HILDA). Section \ref{transparency-qld} then demonstrates and discusses some possible approaches including XAI, IAI and refutation tests which can all offer some insight into the causal effects and therefore help inform policy decisions.

\section{A brief introduction to causal machine learning} \label{sec:introducing-causal-ml} 
What fundamentally separates causal machine learning from the more typically discussed predictive machine learning is that the latter is concerned with predicting outcomes while the former is concerned with predicting treatment effects. The standard definition of a treatment effect in econometrics relies on the Potential Outcomes (PO) Framework \citep{Imbens2015Causal}. Here, in the binary case a treatment effect ($\tau_i$) is the difference between the potential outcome as a function of treatment status $Y_i(W_i)$ where a unit receives the treatment and the potential outcome where a unit does not $$\tau_i=Y_i (1)-Y_i (0).$$

There is an obvious problem here, that one cannot both treat and not treat a unit at a given point in time so in effect, we have to impute counterfactual potential outcomes to do causal inference. This is called the ‘Fundamental Problem of Causal Inference’ \citep{Holland1986Statistics}. It means that unlike for predictive machine learning, in real world data we lack ground-truth treatment effects on which to train a model. It also means that we are relying on a series of causal assumptions the two key ones being the Stable Unit Treatment Value Assumption (SUTVA), and the Independence Assumption \citep{Imbens2015Causal}. The Independence Assumption is required for a causal effect to be considered identified. Essentially it means assuming that treatment assignment is exogenous (as in an experiment), partially exogenous (as in an instrumental variables approach) or endogenous but we will model out the endogeneity for example with a set of control variables (as in control-on-observables) or additional assumptions (as in a difference-in-differences design).

In parametric modelling --- given identifying assumptions hold and a linear parameterisation of the relationship is appropriate --- it is easy to model causal effects by fitting outcomes, in causal machine learning, predictive methods need to be adapted as regularisation shrinks the estimated effect of individual variables towards zero \citep{Chernozhukov2018Double/debiased}. This can be achieved either through specific methods designed to give asymptotically unbiased causal estimates e.g. the causal tree \citep{Athey2016Recursive} or generic estimators designed to plug in estimates from arbitrary machine learning methods e.g. meta-learners \citep{Künzel2019Metalearners,Nie2020Quasi-Oracle}. In all these cases though, the methods still do not have access to ground-truth treatment effects and still require SUTVA and independence assumptions meaning that the exercise is not simply one of maximising fit on held-out data.

Causal machine learning is a broad term for several different families of methods which all draw inspiration from machine learning literature in computer science. The most widely-used method here and our focus for this paper is the causal forest \citep{Wager2018Estimation,Athey2019Generalized} which uses a random forest made up of debiased decision trees to minimise the R-loss objective \citep{Nie2020Quasi-Oracle} in order to estimate HTEs (generally after double machine learning is applied for local centering). The causal forest (at least as implemented in the generalised random forest paper and companion R package \textit{grf}) consists of three key parts, local centering, finding kernel weights and then plug-in estimation. Local centering removes selection effects in the data (assuming we meet the assumptions of control-on-observables identification) by estimating nuisance parameters using two nuisance models one estimating treatment assignment, one estimating outcomes both as a function of a set of control variables \citep{Athey2019Generalized}.\footnote{Note that local centering is not always strictly neccessary. Many causal forest studies use experimental data for example \citet{ajzenman_behavioral_2022,zhou_targeting_2023} and so do not require local centering \citep{Wager2018Estimation} However, in practice papers written after \citet{Athey2019Generalized} which added local centering to the causal forest generally use it. This may simply be for reasons of simplicity (nuisance models are estimated automatically anyway) or because it may improve the efficiency of the estimator per \citet{abadie_large_2006}. For this reason while papers using experimental data do not include explicit identification through nuisance models per se, in practical terms the process of estimation is identical and so the points made in this paper around estimation of effects in observational data are entirely applicable to cases where experimental data are used as well.} The term nuisance here means that the parameters themselves are not the target of the analysis, but are neccessary for estimation of the actual quantity of interest, a treatment effect. This local centering is an adaptation of the double machine learning method which is a popular approach to average treatment effect estimation \citep{Chernozhukov2018Double/debiased}. These models can use arbitrary machine learning methods so long as predictions are not made on data used to train the nuisance model (this is in order to meet regularity conditions in semi-parametric estimation \citep{Chernozhukov2018Double/debiased}). In practice, in the causal forest, nuisance models are generally random forests and predictions are simply made out-of-bag i.e. only trees for which a datapoint was not sampled into its training data are used to make predictions. The residual from these predictions are taken to be the locally centered data. This locally centered data is then fed into a final model designed to find heterogeneity in the data by minimising R-Loss \citep{Nie2020Quasi-Oracle}. Predictions are not made directly out of this model as with a standard random forest, instead this forest is used to derive an adaptive kernel functionn to define the bandwidth used in CATE estimates. Essentially, this weight is based on how many times for a given covariate set $x$, each datapoint in the sample falls into the same leaf on a tree in the ensemble as a datapoint with covariate values $x$. These weightings are then used in a plug-in estimator (by default Augmented Inverse Propensity Weighting) to obtain a final CATE estimate. This is essentially just a weighted average of doubly robust scores with weightings given by the kernel distance according to the forest model. More formally for CATE estimate $\hat{\tau}(x)$, kernel function (from the final causal forest model) $K(\cdot)$ and doubly robust scores $\hat{\Gamma}$ $$\hat{\tau}(x) = \frac{1}{n} \sum_{i=1}^{n} K(X_i - x) \cdot \hat{\Gamma}_i
$$where doubly robust scores are estimated using the nuisance models for outcome ($m(\cdot)$) and treatment assignment ($e(\cdot)$) .$$\hat{\Gamma}_i = \left( \frac{W_i Y_i}{\hat{e}(X_i)} - \frac{(1 - W_i) Y_i}{1 - \hat{e}(X_i)} \right) + \left( \hat{m}_1(X_i) - \hat{m}_0(X_i) \right) - \left( \frac{W_i - \hat{e}(X_i)}{\hat{e}(X_i)(1 - \hat{e}(X_i))} \right) \left( \hat{m}_1(X_i) - \hat{m}_0(X_i) \right).$$

There are also other approaches such as generic methods with R-Learner \citep{Nie2020Quasi-Oracle,Semenova2021Debiased}, single causal trees \citep{Athey2016Recursive} causal Bayesian Additive Regression Trees and Bayesian Causal Forest \citep{Hahn2020Bayesian}, other meta-learners like X-Learner \citep{Künzel2019Metalearners}, DR-Learner , and optimal treatment rule SuperLearner \citep{van_der_laan_super_2007}. While some of this paper is specific to the causal forest, most of the problems discussed here and some of the solutions proposed should be applicable to any causal machine learning approach estimating heterogeneous treatment effects.

The reason for considering all these approaches together is that the collective labelling of them as causal machine learning tells us something about how they are likely to be used in practice --- and the challenges they might present. They are cutting-edge methods that are relatively new in policy research and so there is not much existing expertise in their use. They present new possibilities in automating the selection of models, removing many of the model-design decisions that a human researcher makes and the assumptions that come with these decisions but also rely on black-box models in a way traditional explanatory models do not \citep{Breiman2001Statistical}.

An offshoot of causal learning is what we will term 'prescriptive analysis'. This uses causal models but treats them in a predictive way to make automated decisions. For example, learning decision rules from causal inference \citep{manski_statistical_2004} is a good example of this and approaches to fitting models from HTE learners are already well established (e.g. \citealt{Athey2020Policy,Zhou2018Offline}). However, simply using a causal forest to assign treatment based on the treatment which maximises expected outcome would also be an example of prescriptive analysis, even though the model itself is a causal model that could be used for causal analysis as well. The prescriptive model is a special case as the peculiarities of it as a model that in some way sits between a purely predictive and purely causal model merit special attention. It is not something that is currently being used in policy-making, to our mind it is not a desirable aim nor is it one we treat as a serious policy-making process. However, when talking about join decision-making with a human being it will be useful to have this extreme case as one extreme in the domain where all decision-making power is given to the algorithm.

\section{The rationale for heterogeneous treatment effect estimation with causal machine learning in public policy} \label{sec:rationale}
It is worth briefly pausing to discuss why we might want to use these novel methods for policy evaluation at all. This is particularly important because to the best of our knowledge, causal machine learning has not actually been used in a policy-making process yet. This section presents the current status of heterogeneous treatment effect learning in policy analysis and argues that these methods can fit nicely into an evidence-based policy framework.

While analysis to inform policymaking has been an explicit focus in the methods literature \citep{Lechner2023Causal}, most of the interest in using these methods for policy evaluation so far have come from academic researchers. It is hard to know whether these methods have been used in government or if these academic publications have been used to inform decision-making. As analysis for public policymaking within or in-partnership with government is often not published it is difficult to identify cases where causal machine learning has directly affected decision-making. We are aware of at least one case where it was used by government – a partnership between the Australian Capital Territory Education Directorate and academic researchers to estimate the effect of student wellbeing in ACT high schools on later academic success \citep{Cárdenas2022Youth}. There is however a much larger body of policy evaluation conducted by academic researchers which could be used in policy decisions, but there is no evidence that they have been used in this way (e.g. \citealt{Kreif2021Estimating,Cockx2022Priority,Chernozhukov2021Causal,Tiffin2019Machine}).

Is there value to be obtained from using the causal forest then? We see use of the causal forest as slotting nicely into an evidence-based policy framework where there is a history of porting over causal inference tools from academic research to help improve public policy \citep{althaus_australian_2018}. Of course policy is still incredibly under-evaluated (for example in the UK a \citet{national_audit_office_evaluating_2021} report found 8\% of spending was robustly evaluated with 64\% of spending not evaluated at all) but some of these tools have proved very useful at least in areas of government culturally open to such policy approaches. Being able to identify who is best served by a program and who is not could be knowledge that is just as important as an overall estimate of average effect. Being able to do so in a flexible way, with large datasets is well suited to government. 

In policy evaluation problems, often theoretical frameworks, particularly around treatment effect heterogeneity are quite poor when compared to academic research \citep{levin-rozalis_abduction_2000}. The reason for this is that evaluation are run for pragmatic reasons (because someone decided in the past that the program should exist for some reason), not because the program sits on top of a body of theory that allows for a very robust theoretical framework. Without a strong theoretical framework, there can be little justification for parametric assumptions around interaction effects or pre-treatment specification of drivers for these effects. In addition, the specifics of particular programs often defy the theoretical expectations of their designers because of the specificities of individual programs \citep{levin-rozalis_abduction_2000}. This context makes data-driven exploration of treatment effect heterogeneity particularly attractive because ex ante hypotheses on treatment effect heterogeneity are not needed. In addition, evaluators often have access to large, administrative datasets that can be particularly useful in machine learning methods. The estimation of heterogeneous treatment effects is useful for several reasons. It may help researchers to understand whether a program that is beneficial on average will close or widen existing gaps in outcomes (or even harm some subgroups) and how well findings will generalise to different populations \citep{cintron_heterogeneous_2022}. It can also help to understand moderators that may be pertinent to program design decisions \citep{zheng_estimating_2023}. For example, a program to encourage vaccination in an English-speaking country that has a low treatment effect for non-English-speakers may need to look into building in additional outreach to these communities to improve outcomes.

Our intention in this paper is not to lay out a grand vision for a policy process policy informed by HTE learners. We also do not mean to argue that causal forest or other HTE learners will be able to overcome cultural barriers to adoption within government, rather we make two strictly normative contentions: that there is value to using these methods for policy evaluation in some cases and that addressing transparency challenges is necessary to allow these methods to add value in a policy process. Doing so will improve the usefulness of these tools to policy-makers and address justice issues for those subject to policy.

\section{Transparency and causal machine learning} \label{sec:transparency-needs}
\subsection{The issue of transparency in causal machine learning is similar to that in predictive applications…}
While there is little existing literature on transparent causal machine learning, we can borrow from the much larger literature on transparency in predictive machine learning. We can draw from the predictive literature in laying out a definition of transparency, why it’s desirable, and use it to help find solutions to transparency problems. For the purposes of understanding models and for the purposes of oversight, the concerns are similar. These models are still black-boxes, they are still informing decision-making and in the case of democratic governments making these decisions, there are still expectations around accountability.

\subsubsection{Defining transparency}
When governments employ machine learning tools, there is arguably an obligation that members of the public have a degree of transparency that is not the case for most private sector uses. In some jurisdictions, versions of this obligation have been passed into law (most prominently the EU’s ‘right to explanation’ regulations \citep{Goodman2016European}). Even where transparency isn’t enshrined in law, we will make the assumption which most of the rest of the literature makes that transparency is good and to some degree necessary when using machine learning in government \citep{IreniSaban2022Ethical}. The nature of this need is unclear though and transparency here is not actually one concern, but a range of different, related concerns. Importantly, this critical AI literature is largely about predictive models used in policy implementation

This paper draws on the \citet{Mittelstadt2016ethics} survey of the ethical issues with algorithmic decision-making to map out these transparency issues. Particularly important for the public are what that paper calls unfair outcomes, transformative effects and traceability. The former two (the “normative concerns”) have a direct effect on outcomes for members of the public whether through algorithms that discriminate in ways we judge morally wrong, or by the very use of these algorithms changing how government works (e.g. erroding the standards of transparency expected).

Transparency is not just about understanding models though, it is also about holding human beings accountable for the consequences of these models, what \citet{Mittelstadt2016ethics} calls traceability. Traceability is a necessary part of a process of accountability. Accountability can be seen as a multi-stage process consisting of providing information for investigation, providing an explanation or justification and facing consequences if needed \citep{Olsen2017Democratic}. The problem with accountability for machine learning systems is clear, they can obfuscate the exact nature of the failure, make it very difficult to obtain an explanation or justification. It can be difficult to know who should face consequences for problems or whether there should be consequences at all. Was anyone negligent, or was this more or less an unforeseeable situation (for example, the distribution of new data has shifted suddenly and unexpectedly) \citep{SantonideSio2021Four,Matthias2004responsibility}? This means that models need to be well enough explained that policy-makers can understand them enough to be held accountable for the decision to use them. It also means that causal machine learning systems and the chains of responsibility for these systems need to be clear enough that responsibility can be traced from a mistake inside the model, to a human decision-maker. Finally, it also means that in cases where traceability is not possible due to the complexity of the analysis --- a so called ‘responsibility gap’ --- such analysis should only be used if the benefits somehow outweigh this serious drawback \citep{Matthias2004responsibility}. In the worst case scenario, this opaqueness could not only be an unfortunate side-effect of black-box models, but an intended effect, where complex methods are intentionally used to avoid responsibility for unpopular decisions \citep{Zarsky2016Trouble,Mittelstadt2016ethics}.

An extreme case where a responsibility gap is possible, one that occurs commonly in the predictive literature is what we term prescriptive analysis. Here the machine learning model is directly making decisions without a human in-the-loop. As far as we are aware, no public policy decisions are being made based on causal estimates (analogous to the kinds of automated decisions firms entrust to uplift models when for example targeting customers with discounts). However, even with a human in the loop, there can still be a responsibility gap where the human fails to perfectly understand the fitting and prediction procedures for a model. We can draw on the predictive literature to help solve this problem. When it comes to prescriptive analysis the issues are very similar to those in the predictive literature where there is history of direct decision making by AI models \citep{IreniSaban2022Ethical}. On the other hand, when a human is in the loop on the decision like in explanatory causal analysis where the model is a tool to help understand the drivers of treatment effect heterogeneity, there is less existing theory to draw on. It can best be seen as a kind of human-in-the-loop decision where the human is given a relatively large amount of information meaning that we need to understand where the human decision-making responsibility and that of the algorithm exist distinctly and where we cannot disentangle them \citep{Busuioc2021Accountable}. In the latter case, it will be important for practitioners to construct processes that still allow for accountability (such as along the lines of \citet{Olsen2017Democratic}). An important part of this will be making sure that governments know enough about the models they are using to be held accountable for these joint decisions. It is also important to recognise that there are likely to be responsibility gaps that would not exist with simpler methods \citep{SantonideSio2021Four,Olsen2017Democratic}. This is an unpleasant prospect and these gaps must be minimised. Ultimately, new norms may have to be built up over time about how to use this technology responsibly and hold governments accountable for their performance just as norms and lines of accountability are still forming for predictive machine learning applications \citep{Busuioc2021Accountable}. Governments should also be aware of these downsides before putting causal machine learning methods into practice.

\subsubsection{Methods for achieving transparency}
The solutions we might employ to help understand causal models are relatively similar to those in the predictive literature. This is because the underlying models are generally identical (e.g. metalearners \citep{Künzel2019Metalearners}) or at least very close to existing supervised machine learning techniques (e.g. causal forest). This means that many off the shelf approaches need little or no modification to work with causal models. 
The two families of solutions we can use are both drawn from the predictive AI literature, they are explainable AI (XAI) and interpretable AI (IAI). XAI uses a secondary model to give a local explanation of a black box algorithm, the advantage of this is that a user gets some amount of explanation while not lessening the predictive power of the black-box. Some examples of common XAI approaches are LIME (Local Interpretable Model-agnostic Explanations) \citep{Ribeiro2016Model-agnostic} which perturbs data in small ways then fits a linear model on the outcomes of predictions made with perturbed data to create local explanations or SHAP (SHapley Additive exPlanation) \citep{lundberg_unified_2017} which uses game theory modelling and retraining of models with different sets of covariates to partial out the effect that variables have on predictions. In contrast, IAI approaches give global explanations but at the expense of limiting model selection to ‘white-box’ models that are usually less powerful than typical black-boxes \citep{Rudin2019Stop}. An example of a ‘white-box’ model is a decision tree; here for a given data point one can trace a path through the decision tree that explains exactly how a decision was reached. There are approaches other than just fitting a white-box model initially, for example, several approaches have been proposed to simplify a black-box model to a decision tree by leveraging the black-box model to improve fit over simply fitting a decision tree on the training data directly \citep{Domingos1997Knowledge,liu_learning_2014,frosst_distilling_2017,sagi_explainable_2020}.

Causal machine learning already commonly employs some elements of both the XAI and IAI toolkits, for example, the variable importance metrics or SHAP values presented as outcomes of causal forest analysis could be seen as XAI efforts to explain the individual causal estimates \citep{Athey2019Generalized,Tiffin2019Machine,Kristjanpoller2023Determining}. On the other hand, single causal trees are an interpretable way of estimating the heterogeneous treatment effects \citep{O'Neill2018Causal} and policy allocation rules are a good way to extract insights from black-box HTE models \citep{Sverdrup2020policytree:}. \citet{Athey2019Estimating} graph treatment effects across variables using quantile splits. However, these limited approaches aim to understand the models in specific ways but do not amount to an approach emphasising model transparency, particularly not for oversight purposes. Some basic tools then are already in use and given the structural similarity between causal and predictive models, still others can likely be adapted to improve model transparency.

\subsection{…but there are also important differences from predictive ML}
There are some key differences between the predictive and causal cases for model transparency.  The three main ones are the lack of ground truth in causal inference \citep{Imbens2015Causal}, the role of nuisance model and human understanding in applying the analysis to real-world applications. On the first point, lacking causal inference detaches causal machine learning from the hyper-empirical world of predictive modelling where there is lots of data and few assumptions \citep{pearl_bayesianism_2001}. In causal inference we need to rely on theoretical guarantees, for example that an estimator is asymptotically unbiased, converges on the true value at a certain speed ($\sqrt{n}$ consistency for models directly estimating effect and $\sqrt[4]{n}$ consistency for nuisance functions) and that it has an error distribution we can estimate. This point will not be a focus of this paper, but it ultimately underpins the more practical differences that are. 

On the second point, causal machine learning presents technical challenges because one generally needs to understand a series of nuisance and causal estimating models and how they interact. Poor estimation of this parameter can result in bias to causal estimates \citep{Chernozhukov2018Double/debiased}. As nuisance parameters it is not necessary to interpret the output to answer the research question, but it is necessary to be able to identify whether the estimates are good estimates or not. 

In predictive learning decisions are often made on the basis of predictions automatically while in causal applications, the estimates generally need to be interpreted by a human being. Following on from this, in general predictive systems are used for individual-level decisions (e.g. targeting product recommendations) while the nature of causal questions, particularly in government means that we are interested in outcomes across an entire system (e.g. would changing the school-leaving age boost incomes later in life). Governments generally do not have the capacity (or mandate) to apply policies at the individual-level in many policy areas even if it is in theory possible to do such a thing with individual-level treatment effect estimates. For this reason, there is similar or somewhat less importance in having model transparency for oversight in the causal case compared to the predictive one, but there is the same need for oversight over what we argue is a joint decision made by the human policy-maker and the machine learning system \citep{Citron2007Technological,Busuioc2021Accountable}. For the same reason, it is also important that there is some transparency in the machine learning system for decision-makers and analysts who have to extract insight from the analysis, critique the modelling and weight how much they trust the evidence.

\subsubsection{Models are structured differently}
The most rudimentary difference in the structure of models is that causal machine learning methods generally involve the fitting of several models with different purposes where predictive applications typically involve fitting one, or several with the same purpose in an ensemble method \citep{Chernozhukov2018Double/debiased,Athey2019Generalized}. For example, in the case of DML-based methods (including the casual forest), this involves fitting two nuisance models and then employing some other estimator to generate a treatment effect estimate from the residuals of these models.

The transparency needs for these two kinds of models varies. One can imagine research questions where it is helpful to understand the nuisance models as well as the final model, but for the most part this is not neccessary. We still need some amount of transparency over nuisance functions, mostly to diagnose problems in model specification. The goal of nuisance modelling is not to maximise predictive power and try and get as close to the Bayes error as possible, rather it is to model the selection effects out of treatment and outcome \citep{Chernozhukov2018Double/debiased}. Checking the distribution of nuisance parameters is useful here, how well do models fit the data? How well is the overlap assumption met? There are also a range of non-parametric refutation tests to check how well a given set of nuisance models \citep{Sharma2021DoWhy:}.

Another problem this raises is that explaining or making a model interpretable can only explain the functioning of that one model, but sheds little light on the effect this model has on (or in conjunction with) the other models. Some generic models could trace effect through the whole pipeline of models (for example LIME). However, in this case, it would not be possible to separate out whether the explanations pertained to orthogonalisation or effect estimation. While tools designed for predictive models can be helpful, tools specifically made for causal modelling that account for a series of models which each have different objectives would be even more useful.

\subsubsection{Transparency is more important to users as understanding the model can lead to causal knowledge}
Unlike in predictive applications where transparency is often an orthogonal concern to the main objective of the model (i.e., predictive accuracy), in causal applications, a model is more useful to users when they can understand more of the model structure because the purpose of a model is to inform human beings. Causal machine learning is generally concerned with telling the user something about the data-generating process for a given dataset (some kind of treatment effect) so it can be useful to provide model transparency to suggest patterns in the data even if these are not actually being hypothesis tested. For example, \citet{O'Neill2018Causal} use an interpretable causal tree to provide some clustering to roughly explain the treatment effects in their causal forest. \citet{Tiffin2019Machine} uses SHAP values to lay out possible drivers of treatment effect heterogeneity in a study of the causes of financial crises. SHAP values are calculated by looking through all the combinations of variables seeing how predictions change with a variable included versus when it is excluded. The average marginal effect of each variable is taken to be its local effect \citep{lundberg_unified_2017}.

When trying to build a theory of transparency then, the philosophical basis of the critical predictive literature which focuses on questions of power, ethics and information asymmetries between the user and the subjects of algorithms misses usability --- the role of transparency in explaining causal effects to help inform decisions. This need means we need to see these tools through more of a management theory lens, looking at how to get the best possible information to decision-makers for a given model. The key problem here is one of trust and transparency. Can we give users the tools such that they can perform analysis that reflects real-world data-generating processes? Can we also make sure they understand the model well enough to work well in collaboration with it, that is to weight its evidence correctly and not underweight (mistrust) or overweight (naively trust) its findings just because it is an inscrutable black box?

There is unfortunately only a little literature in the field of decision science which asks how human beings incorporate evidence from machine learning sources sources into their decision-making \citep{Logg2019Algorithm,Green2019Disparate}. The risk here is that humans either irrationally trust or mistrust the algorithm because they do not understand it and this can lead to poor outcomes \citep{Busuioc2021Accountable,GurAli2022Targeting}. This effect is often called automation bias. One could reasonably assume that causal machine learning algorithms given their complexity and their novelty might cause a more potent biasing effect than traditional regression approaches which are more familiar to those doing causal inference \citep{Breiman2001Statistical,Imbens2021Breiman's}. \citet{Logg2022psychology} explains this effect as being a result of human beings having a poor ‘Theory of Machine’, the algorithmic analogue of the ‘Theory of Mind’ by which we use our understanding of the human mind to assess how a human source of evidence reached the conclusion they did and whether we should trust them. When it comes to algorithms, Logg argues that decision-makers often over-weight this advice as they do not understand what is going on inside the algorithm but instead see it as incomprehensible advanced technology that seems powerful and objective. \citet{Green2019Disparate} concur as their participants showed little ability to evaluate their algorithm’s performance even when trusting it to make decisions that were obviously racially biased. 

A good decision-maker using an algorithmic source of evidence needs enough understanding to be able to interrogate evidence from that source and the process that generated it, just as a good decision-maker relying on human sources of evidence will know what questions to ask to verify this information is worth using \citep{Busuioc2021Accountable}. Having a good Theory of Machine for a causal machine learning model then means needing to understand the final model, but it also means understanding the algorithm that gave rise to the model \citep{Logg2022psychology}. An analyst needs to be able to challenge every step of the process from data to estimate, a decision-maker needs a good enough understanding to provide an outside eye in case the analyst has missed any flaws and to be able to decide how much weight the evidence should be given \citep{Busuioc2021Accountable}. This means that we should aim for what \citet{Lipton2018Mythos} calls algorithmic transparency (i.e. understanding of the fitting algorithm) to the extent it is possible as well as just model transparency.

In cases where causal machine learning is being used for orthogonalisation --- that is meeting the independence assumption by controlling for variation in outcome that is not orthogonal to treatment assignment (e.g. DML and methods derived from it) --- there is an additional need not for transparency in the traditional sense, but rather to understand a model well enough to diagnose problems with identification \citep{Sharma2021DoWhy:}. For example, it might be important that the approach to identification used by the nuisance functions makes sense to a domain expert. Of course, it might be possible that the model is drawing upon relationships in data that are legitimate for identification, but that the domain expert cannot comprehend, but there are processes by which we can iterate on and test such models. For example, \citet{GurAli2022Targeting} lays out a procedure for iteratively constructing an interpretable model of HTEs based on an XAI output from a causal machine learning model. The ‘transparency’ that is useful here does not just come from transparency tools designed for the predictive world though. Other causal inference diagnostics can be brought in as what are effectively AI transparency tools solving problems of identification. For example, refutation tests like Placebo Treatment or Dummy Outcome tests could be useful in providing algorithmic transparency for ATE estimation (though they are untested when it comes to causal machine learning HTE estimation) \citep{Sharma2021DoWhy:}.

\subsubsection{The distance between causal models and real-world impact is greater because humans are the ultimate decision-makers}
The link between the results of causal analysis and real-world action is also generally less clear than in predictive applications which changes the importance of transparency. Generally, causal analysis is further distanced from making actual decisions than predictive models are. The kinds of questions causal analysis is used to answer (particularly in government) and the complexity of causal identification means that in practice, causal analysis is largely used to inform human decisions by providing a picture of the underlying causal effects rather than driving automated decisions. Of course, predictive applications sometimes involve a human in the loop as well. However in practice, this is rarer in predictive applications (as causal applications almost never lack a human in the loop (see \citealt{rehill_fairness_2023})) and here decision-making it is generally a matter of acting on a single prediction rather than drawing conclusions from an approximation of the whole set of causal relationships in the data (for example in approving loans \citep{Sheikh2020Approach} or making sentencing decisions \citep{Završnik2020Criminal}).

Because there is a human in the loop (who is depending on one’s views, a more trustworthy agent and / or a more impenetrable black-box) drawing on a range of other evidence (or at least common sense), it becomes less important for oversight purposes to have a transparent model. It is of course still useful to be able to scrutinise the human-decision maker and the evidence they relied on to make their decision, but the transparency of the model itself is a less important part of this oversight than it would be were the decision fully automated. Instead the challenge is in understanding a joint decision-making process, one that is not necessarily any less daunting.

As a side-note, this distance sets up the potential for accountability and usability to be adversarially related. Assuming the effect of the causal model on the real-world is always fully mediated through a human's understanding of the model, the usability of the model increases the need for accountability. This is because the human policy-maker can only incorporate evidence that they understand into their decisions so the model needed to understand the evidence used in the decision needs to be complex enough to model that understanding, not necessarily the actual causal forest. For example, if a decision-maker simply made a decision based on a best linear projection (BLP) for a causal forest, the underlying model is essentially irrelevant for accountability because the whole effect is mediated through the BLP\footnote{The BLP regresses the doubly robust scores onto a set of covariates in order to get the best linear model to explain treatment effect heterogeneity. This provides an interpretable model with hypothesis testing that is easy for any reader with experience in linear regression to understand. However, it will not capture non-linear relationships in the data.}. One only needs to understand the BLP to ensure accountability. On the other hand, if decisions are made based on a detailed understanding of nonlinear relationships in the causal forest obtained through powerful usability tools, accountability methods will need to be powerful enough to explain these effects.

There are some key differences between causal and predictive machine learning methods. The nature of models and the way they are likely to be used in practice means that there is still some work to be done in developing transparency approaches specifically for these methods. The following section tries to do this by working through a case study showing some of the possibilities and some of the limitations of the tools that currently exist.

\section{Introducing the returns on education case-study} \label{sec:case-study}
\subsection{Background} \label{sec:bg}
In Australia there has been a great deal of research on the returns on tertiary education, particularly as the rate of attainment of a tertiary qualification grows \citep{Leigh2008Estimating}. The case study in this section and the next will attempt to estimate the causal effect of a marginal year of education in Australia replicating \citet{Leigh2008Estimating} with fully observational methods. We will do this by analysing data from the Household Income and Labour Dynamics in Australia (HILDA) survey from 2021-22 (Wave 21).

We take a fully observational approach to this research controlling for a matrix of pre-treatment variables. This is not the ideal approach for unbiased estimation (per \citealt{Leigh2008Estimating}) but control-on-observables studies with the causal forest are far more common than quasi-experimental designs. Perhaps this is due to a lack of good tooling for quasi-experimental approaches, perhaps it is because local estimates are less useful for CATEs than they are ATE estimation due to sample size and concerns about covering parts of the covariate distribution for CATE estimation that might be ignored in a LATE estimate. It may simply be due to greater confidence in the ability of regression forests to model out selection effects in local centering in a way standard methods cannot.

We fit all models with an ensemble of 50,000 trees on 15360 cases. We identified 34 valid pre-treatment variables which are (listed in Appendix A) to use for fitting nuisance functions and the main causal forest. Some of these variables are strictly speaking nominal, but have some kind of ordering in their coding so have been included as quasi-ordinal variables (country coding which roughly speaking measures cultural and linguistic diversity, occupational coding which roughly speaking goes from managerial to low-skilled). As the causal forest can non-linearly fit this data, it can find useful cut points in this data or simply ignore the variable if it is not useful.

These variables were chosen out of almost 6993 possible controls in the dataset because when trying to orthogonalize we can only use pre-treatment variables \citep{Chernozhukov2018Double/debiased, hunermund_causal_2021} and the variables could be considered post-treatment because present income is being measured in most cases years after the respondent was last in education.\footnote{While there is some missing data, for the purposes of a case study rather than an actual study into the effect we will assume this is missing completely at random (MCAR) \citep{Rubin1976Inference}. While it is unwise to assume this data is MCAR, median imputation is suitable for an application where we are mostly interested in demonstrating the method. To the best of our knowledge, there is no work on the effect that median imputation might have on causal forest estimates. There is the potential for the models to react to these imputed values in way that classical methods would not, for example essentially learning the missingness of data with by the imputed value, but ultimately accurate inference is not the priority in this study and data is relatively complete so simple imputation methods are suitable.} After estimating nuisance models for treatment value propensity $e^*(\cdot)$ by regressing treatment $W$ on the covariates $X$ with a regression forest and outcome $m^*(\cdot)$ by regressing outcome $Y$ on $X$ again with a regression forest, we can then estimate $\widetilde{\tau}\left(\cdot\right)$ as a standard prediction problem minimising R-loss as a function of $X$ \citep{Nie2020Quasi-Oracle}. Here $\tau(\cdot)$ are candidate heterogeneity models that try to explain heterogeneity after local centering. $\Lambda_n$ is a regulariser, here regularisation implicit and provided by the structure of the ensemble and trees not least by enforcing out-of-bag fitting and honesty.

$$\widetilde{\tau}\left(\cdot\right)=argmin_\tau\left(\frac{1}{n}\sum_{i=1}^{n}\left[\left\{Y_i-m^\ast\left(X_{IDi}\right)\right\}-\left\{W_i-e^\ast\left(X_{IDi}\right)\right\}\tau\left(X_i\right)\right]^2+\Lambda_n\{\tau\left(\cdot\right)\}\right)$$
 
It is also worth noting that this approach is not necessarily the ideal one for seriously studying this question empirically. A fully observational study is much less convincing than one employing the well established instrument \citep{Leigh2008Estimating}. Even if it were convincing, there are likely key confounders that are missing from this dataset which focuses heavily on family background. This is because variables have to be pre-treatment but data has to be collected at a point in time where there is income data, i.e. much later. 

Finally, it may be useful to include certain post-treatment variables in the heterogeneity model but not in the nuisance models (for example the number of children someone has had). These cannot be controls because they are post-treatment but could be important moderators \citep{pearl_causality_2009}. For example, women who have children after their education tend to have lower incomes than similar women who did not have children and therefore lower returns on education \citep{cukrowska-torzewska_motherhood_2020}). These would be bad controls in the nuisance models \citep{hunermund_causal_2021}, but improve fit and help us to uncover the presence of an important motherhood effect in the heterogeneity model \citep{watson_heterogeneous_2023,celli_causal_2022}. However, while grf can handle different sets of variables for different models, this is not the case for the EconML package in Python which we use to generate SHAP value plots (though in Section \ref{sec:xai} we are skeptical of the validity of SHAP values for these models given the different roles of variables in the nuisance and heterogeneity models). For this reason we choose a more limited model (and suggest EconML change its approach to be more like that of grf).

The causal forest produces an APE estimate of \$5753 per additional year of education with a standard error of \$316. The real value of the method though is of course in analysing CATEs which we will do through XAI and IAI lenses. While neither of these groups of tools actually amount to showing a causal relationship that variables might in driving treatment effect variation, these results are still useful in seeking to understand causal effects.

\section{Transparency in the Queensland case study} \label{transparency-qld}
\subsection{Using XAI tools} \label{sec:xai}
This section considers how the transparency problems might be addressed and what issues may be insurmountable. For the most part the issues of understanding and oversight will be combined as they both encounter similar technical barriers. There are two main XAI approaches that have been proposed for the causal forest, the first is a more classic predictive machine learning approach in SHAP values \citep{lundberg_unified_2017}. The second is a variable importance measure which has somewhat more humble ambitions --- it does not seek to quantify the impact of each variable in each case but instead tries to show which variables are most important in fitting the forest.

\subsubsection{SHAP}

We start by using an XAI approach, in particular the SHAP method which has previously been applied to causal forest analysis \citep{Tiffin2019Machine}. SHAP values decompose predictions into an additive combination of effects from each variable for a local explaination (that is the contribution of each variable is only locally to that part of the covariate space) \citep{lundberg_unified_2017}. SHAP values are based on Shapley values which provide a fair way to portion out a pay-off amongst a number of cooperating players in game theory. It does this by considering how the prediction changes when different sets of features are removed (set to a baseline value) versus when they are included \citep{lundberg_unified_2017}. More details on the calculation of SHAP values can be found in \citet{lundberg_unified_2017}. In this case the pay-off is the difference between the causal forest prediction and the average treatment effect and the players are the different covariates. It uses the predictions of a causal forest in much the same way it would use the predictions of a predictive random forest to generate SHAP explanations.

Unfortunately, good methods to calculate SHAP values exist only for forests implemented in the Python EconML package, not the R grf package. Conversely, the EconML package lacks some of the features of grf and is implemented slightly differently. This accounts for differences in results between these outputs and the grf outputs in other sections. In addition, due to the long time to compute SHAP for a large ensemble, we use an ensemble of just 1000 trees here.

It is worth stating that it is not clear that SHAP values are suitable for this application. There is some question among the maintainers of grf as to whether SHAP values are appropriate for a causal forest given the way the forest is used to construct kernel weights rather than directly estimating based of aggregated predictions \citep{grf-labs_add_2021}. This argument would apply in theory to any predictive XAI tool which does not account for the specific estimation strategy of the generalised random forest estimators \citep{Athey2019Generalized}.

\begin{figure}[!h]
    \caption{Aggregated SHAP plot explaining the HTE estimate across the distribution}
    \centering
    \includegraphics[scale=0.45]{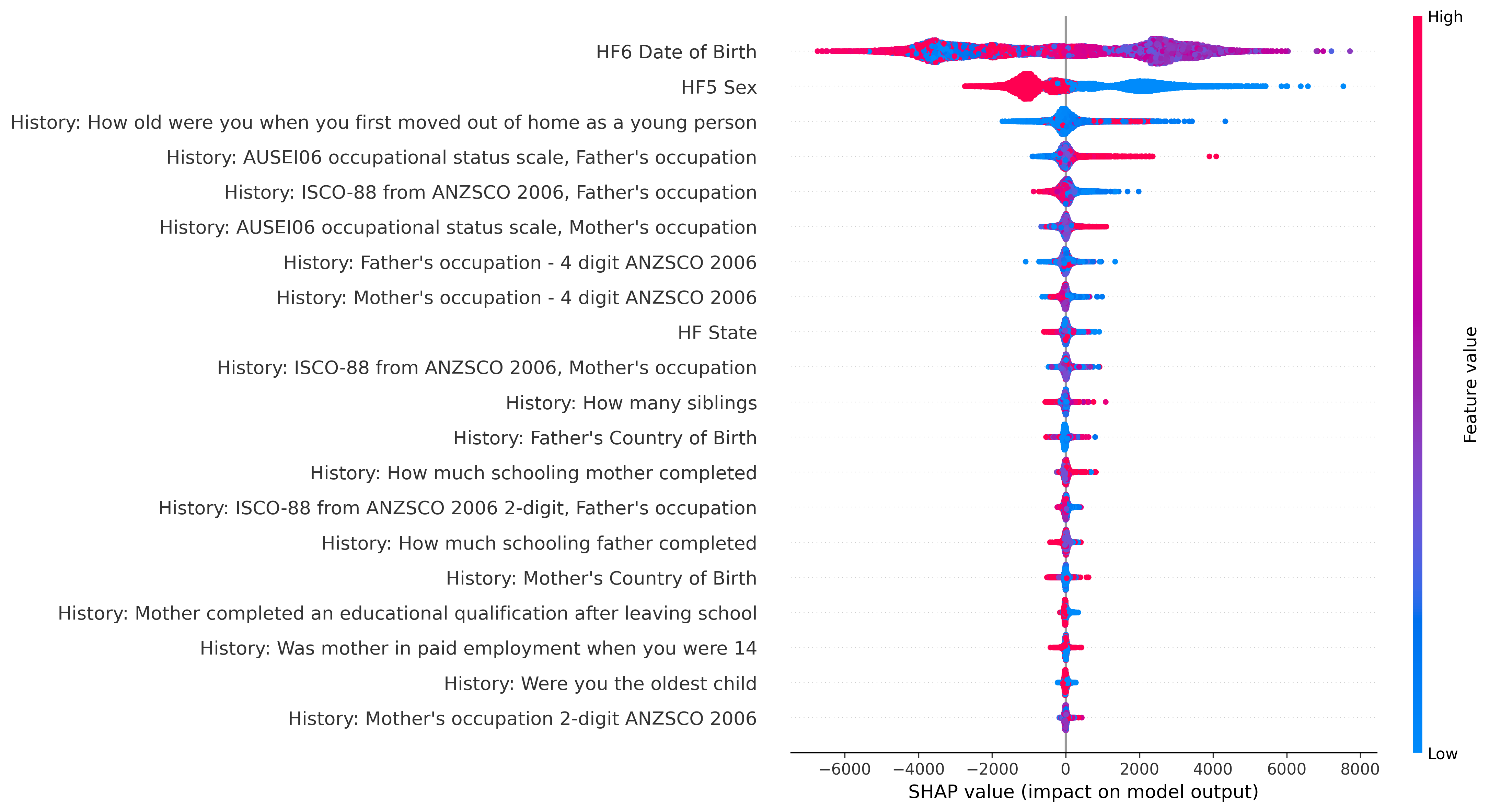}
    \label{fig:fig4}
\end{figure}

\begin{figure}[!h]
    \caption{Waterfall plot explaining the HTE estimate for a random individual}
    \centering
    \includegraphics[scale=0.35]{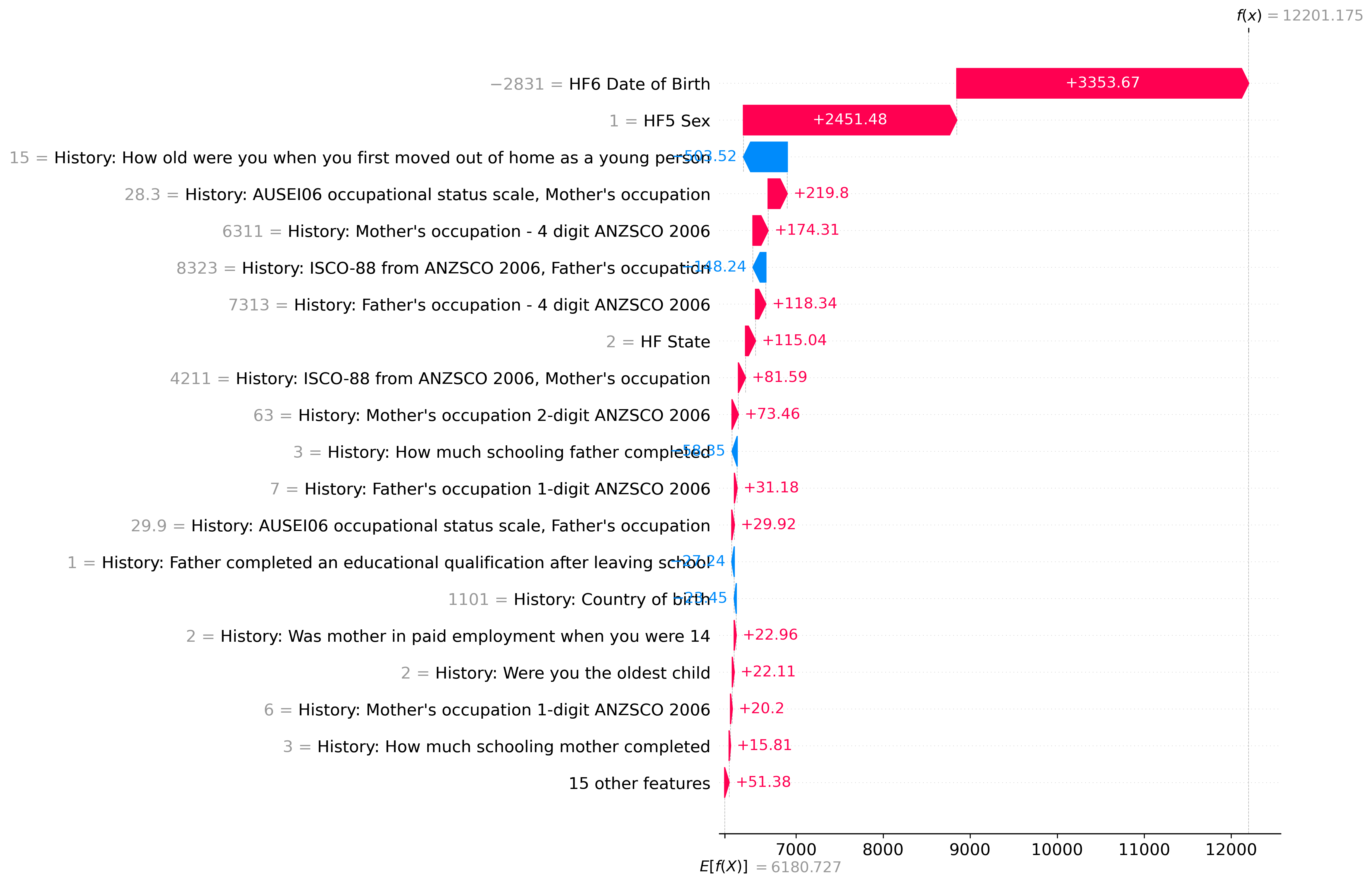}
    \label{fig:waterfall}
\end{figure}

There are two different ways to visualise SHAP, as a set of individual waterfall plots (like Figure \ref{fig:waterfall} explaining specific cases or as an aggregate plot like Figure \ref{fig:fig4} which summarises the variable contributions of the top 20 variables in the EconML model across all respondents. In a waterfall plot, feature names and values for a specific case are shown on the right and the effect that feature has on the CATE is shown as a red (positive) or blue (negative) bar. This deviation is from the average treatment effect. On the aggregate summary plot, the value on CATE estimate is shown on the x-axis and the feature value leading to that estimate is shown on the colour scale (per the bar on the right of the plot) For example we can interpret Figures \ref{fig:fig4} and \ref{fig:waterfall} as follows, starting with the former. The model shows that those with a low response for sex (meaning males in the HILDA coding) boosted treatment effects over those with high values (women). Those who are middle-aged have have higher treatment effects than those who are younger or older, and those whose parents worked in higher educated, higher paid, higher status occupations generally had higher treatment effects. There are many other smaller effects. In Figure \ref{fig:waterfall} we see in this specific case that the individual being middle-aged (this value corresponds with being born in 1963) and a man has pushed up their estimated treatment effect while having left home at 15 lowered it slightly. There are several other factors with relatively small effects. Overall this individual has a CATE almost double the ATE.

A waterfall plot of the SHAP values for one randomly sampled individual in this dataset is presented in Figure \ref{fig:fig4} with several more presented in Appendix B to show the variety between plots. Note that the date of birth variable dob is encoded numerically per Unix time, that is the number of days from the 1st of January 1970. This one shows that the respondent is a man with two children and who has never married. The first of these factors gives a large boost in estimate while the second and third lower the estimate.

While SHAP values might be useful they are by their nature local explanations and so it can be hard to extract insight from them for either usability or accountability.

SHAP values --- assuming their validity with the causal forest --- can be an excellent aid for usability. SHAP values give arguably a more 'causal' insight than simply graphing distributions across variables,\footnote{We use the word "causal" here with hesitation given that these effects are not formal causal estimates. We mean instead that using SHAP is meant to give a sense of how treatment effects might vary with a given covariate \textit{ceteris parabus}. Importantly any discussion of drivers of heterogeneity is not strictly causal as these effects do not remove what we might call higher order selection effects i.e. the effect of selection into drivers of heterogeneity that comes from other variables. For example, we might exogenously vary school-leaving age to experimentally estimate returns on education, but estimates of heterogeneity across occupation, gender and parents' occupation would be correlational as neither the experimental assignment nor the causal forest's local centering step is removing the effect of gender and parents' occupation in selection into one's own occupation. A best linear prediction (see Section \ref{sec:blp}) could be interpreted as doing this under strong assumptions (ignorability of endogeneity for all predictors acting as linear controls for each other and the parametric assumption of linearity)} as it aims to take account of the additional effect of a given variable where our plots of effect distributions simply gives a visual sense of correlation. This can help to understand patterns in causal effects that may have been missed otherwise. SHAP plots can even give a sense of interactions between features when viewing a number of different waterfall plots. For example, if having children hurts women's incomes, particularly the the decade after the birth of a child, we might see age have a different effect on the treatment for women than men. While there might be good information here, this is also a drawback. There is also a lot of information that needs to be processed to help a human decision-maker understand causal effects and make a decision and as all these explanations are local. The user may be simply missing the important patterns and not know it because they are awash in useless data. This is not to mention the information from the nuisance models that could be gained through SHAP analysis as well.

SHAP is intuitively useful for accountability because it lays out variable effects in an easy-to-understand way and can be used to break down effects at the individual level. However, this convenience is somewhat misleading. The problem is that SHAP is a relatively poor approach to accountability because it poorly explains joint decision-making. The amount of data it provides on the underlying models can be simply overwhelming, but it also obscures the core question, how did a human make a decision that had real-world consequences based on this model? What local effects were generalised into evidence? What evidence was interpreted as showing underlying causation? What local effects were ignored?

One benefit of SHAP for both usability and transparency though is that it is well-suited to diagnosis of problems in the model, for example, biasing 'bad controls' would show up among the most impactful variables. Equally, variables that should have a large effect but which are not present among the top variables may suggest errors in data. Finally, the local level explanations which we have previously suggested is a limitation could be useful to individuals trying to find modelling errors for accountability as for example, it could allow an individual to examine SHAP scores in their own case and see if the results match their priors about causal effects in their own case. Exactly how to go about updating modelling approach versus ones priors is a tricky question that is beyond the scope of this paper but would be an interesting avenue for future research. It is worth noting as a final point that SHAP values may be infeasible for larger models and larger datasets. SHAP is relatively time complex and so trying to explain results may prove computationally infeasible \citep{benard_variable_2023}.

\subsubsection{Variable importance in heterogeneous treatment effect estimation}
Variable importance seeks to quantify how impactful each variable was is in a given model. In predictive modelling where the techniques were invented there are several methods for doing this aided by access to ground-truth outcomes. For example, we can sum the decrease in impurity across all splits for a given variable or see how performance suffers by randomly permuting a given feature \citep{saarela_comparison_2021}. In the causal forest context (at least in the \textit{grf} package), we lack ground truth and so have to use more heuristic or computationally complex approaches.

There are two main approaches to variable importance. The first ---  which is the simpler of the two --- is about counting uses of the variable in a causal forest. It was developed for the \textit{grf} package. In this approach, variable importance is measured with a heuristic where the value is a normalised sum of the number of times a variable was split on weighted by the depth at which is appeared (by default, exponential halving by layer) and stopping after a certain number (by default four) to improve performance \citep{Athey2019Generalized}. This is a relatively naive measure (something the package documentation itself admits), the naivete is made necessary by a lack of ground truth which prevents the package from using the more sophisticated approaches that predictive forests tend to rely on \citep{NIPS2013_e3796ae8}. This means a whole rethink of the approach to generating variable importance measures is needed, but finding a more sophisticated approach was outside the scope of the \textit{grf} package which was focused on just laying the groundwork for the generalised random forest approach \citep{Athey2019Generalized}.

Taking this depth weighted split count approach, the variable importance for the forest is shown in Table \ref{tab:var_imp}. We see interestingly that date of birth accounts for more than 50\% of depth weighted splits with the top 10 variables cumulatively making up 87\%. Variable importance clearly gives less information about predictions than the SHAP plots (assuming the validity of applying SHAP to the causal forest), however it does tell us some similar things about the factors that seem to drive heterogeneity in causal effects.

\begin{table}[!htbp] \centering 
  \caption{Variable importance for the predictors in the causal forest} 
  \label{tab:var_imp} 
\begin{tabular}{@{\extracolsep{5pt}} lll} 
\\[-1.8ex]\hline 
\hline \\[-1.8ex] 
Rank & Names & Importance \\ 
\hline \\[-1.8ex] 
1 & HF6 Date of Birth & 0.524 \\ 
2 & HF5 Sex & 0.083 \\ 
3 & History: Country of last school year & 0.071 \\ 
4 & History: AUSEI06 occupational status scale, Father's occupation & 0.047 \\ 
5 & History: Country of birth & 0.039 \\ 
6 & History: ISCO-88 from ANZSCO 2006, Father's occupation & 0.026 \\ 
7 & History: Mother's occupation - 4 digit ANZSCO 2006 & 0.025 \\ 
8 & History: ISCO-88 from ANZSCO 2006, Mother's occupation & 0.023 \\ 
9 & History: AUSEI06 occupational status scale, Mother's occupation & 0.020 \\ 
10 & History: Father's Country of Birth & 0.014 \\ 
11 & History: Father's occupation - 4 digit ANZSCO 2006 & 0.013 \\ 
12 & History: Mother's Country of Birth & 0.013 \\ 
13 & History: Ever had any siblings & 0.012 \\ 
14 & History: How much schooling mother completed & 0.012 \\ 
15 & History: How many siblings & 0.011 \\ 
16 & History: How old were you when you first moved out of home as a young person & 0.009 \\ 
17 & History: Mother's occupation 2-digit ANZSCO 2006 & 0.008 \\ 
18 & History: ISCO-88 from ANZSCO 2006 2-digit, Father's occupation & 0.007 \\ 
19 & History: Country of birth - brief & 0.006 \\ 
20 & HF State & 0.006 \\ 
21 & History: How much schooling father completed & 0.005 \\ 
22 & History: ISCO-88 from ANZSCO 2006 2-digit, Mother's occupation & 0.005 \\ 
23 & History: Was mother in paid employment when you were 14 & 0.004 \\ 
24 & History: Father's occupation 2-digit ANZSCO 2006 & 0.003 \\ 
25 & History: Did your mother and father ever get divorced or separate & 0.003 \\ 
26 & History: Mother completed an educational qualification after leaving school & 0.003 \\ 
27 & History: Were you the oldest child & 0.002 \\ 
28 & History: Mother's occupation 1-digit ANZSCO 2006 & 0.002 \\ 
29 & History: Were you living with both your parents when you were 14 years old & 0.002 \\ 
30 & History: Father completed an educational qualification after leaving school & 0.001 \\ 
31 & History: Was father unemployed for 6 months or more while you were growing up & 0.000 \\ 
32 & History: Father's occupation 1-digit ANZSCO 2006 & 0.000 \\ 
33 & History: Was father in paid employment when you were 14 & 0.000 \\ 
34 & History: Aboriginal or Torres Strait Islander origin & 0.000 \\ 
\hline \\[-1.8ex] 
\end{tabular} 
\end{table} 
 
A more sophisticated version of variable importance is implemented in the \textit{mcf} package \citep{lechner_modified_2019} which uses permutation variable importance to estimate variable importance metrics. It does this by randomly shuffling values for each variable in turn and then predicting out new estimates. The change in the error for those predictions gives a sense of how important a given variable is in the model structure. Another approeach is taken in recent work by \citet{hines_variable_2022} and \citet{benard_variable_2023} which tries to estimate the proportion of total treatment effect variance explained by each variable used to fit the causal forest. These two recent approaches both work by retraining many versions of a causal forest with and without variables and finding what percentage of treatment effect variation is explained when a variable is added in. The lack of existing implementations for continuous treatment effects mean it is not possible to use these in this application, however there seems to us to be no reason these approaches could not be adapted to continuous treatments in future.

Variable importance has humbler ambitions than SHAP and arguably benefits from this when it comes to being of use for transparency. Variable importance provides much less data which in turn means it is less likely to be misinterpreted and less likely to be relied on to actually understand the model rather than be a jumping off point for exploratory analysis. It also provides an arguably more global explanation by simply summarising the structure of the causal forest rather than trying to explain individual estimates.

The question is though, does this more limited ambition help in achieving usability or accountability? On usability, variable importance can be a good tool for exploratory analysis for example in identifying possible drivers of heterogeneity for which heterogeneity can be explored further (for example by graphing the effect or by modelling doubly robust scores with a parametric model). In addition, it can be useful for sense-checking that all variables involved are 'good controls'. Any variable that is going to have a substantial biasing effect ala collider bias will have to be split on in order to have a biasing effect. It should therefore show up in the variable importance calculation. There is obviously more that needs to be done to explain heterogeneity than just counting splits. Variable importance is --- at its best --- the starting point for further analysis for usability.

On accountability, variable importance is not particularly useful because it is such a high-level summary of the model being used. If there are for example unjust outcomes occurring because of the model, it is hard to tell this simply from variable importance. By an unjust outcome we mean that somewhere in the complexity of estimating effects, poor local centering, poor estimation of CATEs, poor communication of CATEs, the model has given decision-makers a false impression about the underlying causal relationships this leads them to make an 'unjust' decisions. While we are happy to defer to the role of decision-makers to decide their own definition of justice, modelling which does not allow the decision-maker to make decisions to better their own definition of this are unjust. This is admittedly a convoluted definition but one which is neccessary given the indirect relationship between model predictions and decisions compared to the more direct relationship in predictive modelling \citep{rehill_fairness_2023}. For example, in predictive contexts it can be useful to see if a sensitive variable (or its correlates) has any effect on predicted outcomes with a model being fair if outcomes are in some way orthogonal to these sensitive variables \citep{mehrabi_survey_2019}. In the causal context on the other hand it can be important to know that marginalised groups have lower predicted treatment effects for example because a linguistic minority cannot access a service in their own language. In fact it would be unjust if modelling failed to reveal this and so a decision-maker not understanding this treatment effect heterogeneity could not act to improve access to the program. Just seeing splitting on certain variables then is not indicative of a particularly unjust model --- yet this is the only level of insight we get from variable importance.

\subsection{Using IAI tools}
We can see that when using XAI tools there is a large amount of information to process, and we cannot get a global understanding of how the model works. We might then want to turn to interpretable models. There are two main ways of doing this. The classic approach to IAI for a random forest is to simplify the ensemble down to a single tree as this does not involve additional assumptions \citep{sagi_explainable_2020}. In the case of the causal forest, another approach that is often taken is to simplify the forest down to a best linear projection of heterogeneity \citep{Athey2019Estimating}. This imposes additional assumptions but provides a model that is interpretable to quantitative researchers and in particular, allows for hypothesis testing.\footnote{It is worth noting that the \textit{mcf} Python package which implements a modified causal forest had some interesting approaches to generating interpretable models. For example, it fits interpretable but non-linear models on estimates from a causal forest and uses k-means clustering ala \citep{cockx_priority_2023} to find clusters. These could be useful methods that could be written on at length, however the \textit{mcf} works differently from a standard causal forest and so will not be explored in-depth in this paper.}

\subsubsection{Extracting a single tree}
While there are lots of individual trees in the causal forest and one could pick one (or several) at random to get a sense of how the forest is operating, there are also more sophisiticated approaches that provide ideally a tree that is better than one just chosen at random. \citet{Wager2018Find} suggests a good way to find a representative tree would be to see which individual tree minimises the R-Loss function of the causal forest. This is not a peer reviewed approach, nor one which has even been written up as a full paper but it represents the best proposal specific to the causal forest that we have. Other approaches attempt to distill the knowledge of the black box forest into a single tree that performs better than any individual member of the ensemble (e.g. in \citealt{sagi_explainable_2020, Domingos1997Knowledge, liu_learning_2014}). However, the problem here is that even the smartest methods for extracting a tree of sufficient simplicity to be interpretable require problems where there is enough redundancy in rules --- the underlying structure can be captured almost as well by a few splits in a single tree as by many splits in a large ensemble --- that the problem can be simplified to the few best rules with relatively little loss in performance. Rather than focusing on methods for extracting the best tree then, we instead look at whether enough redundancy exists in this problem to make simplification to a reasonably good tree feasible, analysis that to the best of our knowledge has not been done for an application of the causal forest.

The exact marginal trade-off of adopting a more interpretable model depends on what \citet{Semenova2019study} calls its Rashomon Curve. The Rashomon Curve graphs the change in performance as a model is simplified in some way, for example by a reduction in the number of trees in a random forest until it becomes a single tree. In Figure 5 we show a Rashomon curve comparing the performance of a causal forest (that is the final heterogeneity model) of 50,000 trees against one of 1, 10, 100, 1000 and 10,000 trees and a tree distilled from the 50,000 tree forest.
\begin{figure}[!h]
    \centering
    \includegraphics[scale=0.7]{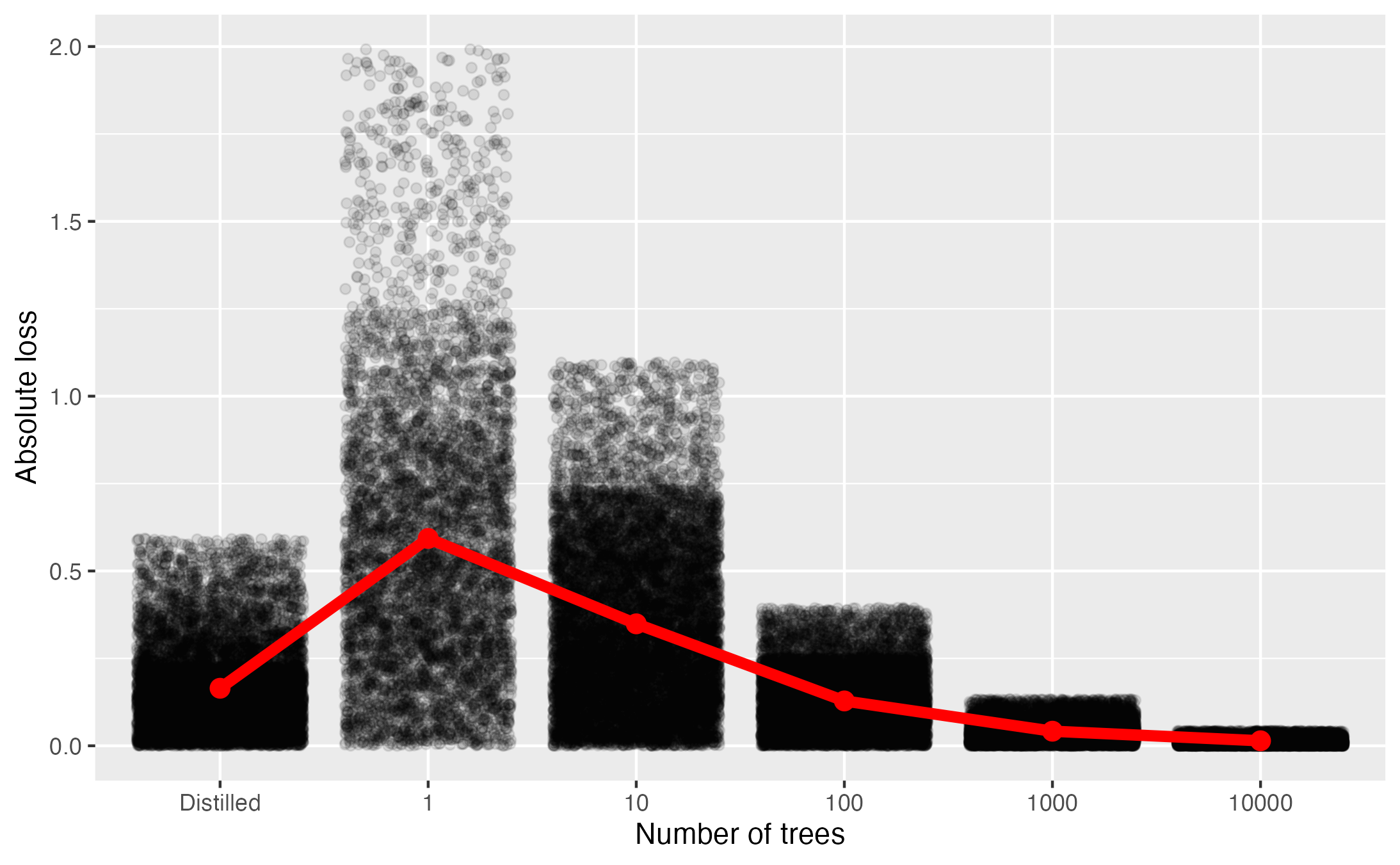}
    \label{fig:fig5}
        \caption{Rashomon curve for the effect of heterogeneity estimating model size (number of trees) on estimates (note, top 2.5\% of values have been trimmed away due to a few very high error predictions making visualisation difficult).}
\end{figure}

In these figures, error is shown as percentage error between the forest and the baseline forest of 50,000 trees. Because we lack ground truth and know that the causal forest is asymptotically unbiased and accuracy should increase as number of trees increases, performance relative to a large forest should give a good indication of performance relative to ground-truth. In all cases, these used identical nuisance functions each fit on 10,000 trees. This reveals substantial loss in accuracy as the size of the forest shrinks, this accuracy is due to an increase in what the \textit{grf} package calls excess error --- error that would shrink towards zero as ensemble size approaches infinity (as opposed to debiased error which is not a function of forest size). The Rashomon curve here may be flatter for problems with lower excess error, however in this case, the trade-off for moving to a single tree seems to be a poor one. To put it in concrete terms, the mean absolute loss for moving from 100,000 trees to a single tree was over 50\% of the comparison value on average.

We can also conceptualise of the trade-off in terms of R-Loss. Table \ref{table:r-loss} shows the results of this comparison with all values subtracted from the R-Loss of the largest ensemble --- 50,000 trees with loss of 3,521,089,910. While R-Loss values are often extremely high and only slightly influenced by the characteristics of a heterogeneity learner \citep{Nie2020Quasi-Oracle} what matters is that lower R-Loss is better. The small difference in R-Loss scores as proportions of this total loss should not be taken as an indication of relative performance or overall poor performance. Note the small, chance increase in performance moving from 50,000 to 10,000 trees is unexpected but not worrying given how small the difference is.

% Table created by stargazer v.5.2.3 by Marek Hlavac, Social Policy Institute. E-mail: marek.hlavac at gmail.com
% Date and time: Fri, Mar 29, 2024 - 01:37:25
\begin{table}[!htbp] \centering 
  \caption{Comparison of R-Loss values for different ensembles relative to the value for 50,000 trees.} 
  \label{table:r-loss} 
\begin{tabular}{@{\extracolsep{5pt}} ll} 
\\[-1.8ex]\hline 
\hline \\[-1.8ex] 
Ensemble size & Relative R-Loss\\ 
\hline \\[-1.8ex] 
50000 & 0 \\ 
10000 & -25,352 \\ 
1000 & 2,494,325 \\ 
100 & 10,409,270 \\ 
10 & 100,362,167 \\ 
1 & 111,189,521 \\ 
Distilled & 10,592,467 \\ 
\hline \\[-1.8ex] 
\end{tabular} 
\end{table} 

An alternative approach is to use distillation to improve performance. Distillation of black-box models to a single tree can improve the performance of a tree over simply fitting that tree on the same data the black-box learner or 'teacher' was fit on. Exactly why this is the case is not yet entirely clear but it is a useful empirical technique \citep{hinton_distilling_2017,dao_knowledge_2021}. The use of distillation with causal forests has recently been proposed by \citet{rehill_causal_2024}. Here we compare the performance of the basic single tree model against a distilled model. We show that the performance of the distilled model is much better than the single tree trained on the raw data and roughly as good as a 100 tree ensemble trained on raw data for both predicting the 50,000 tree ensemble predictions and for minimising R-Loss.

Reducing a model to a single tree would be very useful for both usability and accountability. In both cases, the entirety of the model can be comprehended by people using that model to make decisions and people critiquing the model. The problem is that the trade-off may simply not be worth it. There are already very interpretable methods for quantitative research that are well understood. This is not to mention that there is a much larger suite of methods drawing on the latter approach for removing selection bias that does not involve trying to model out confounding with a relatively simple model (like a linear model or decision tree). For example, a practitioner could use instrumental variable regression, difference-in-differences etcetera. So even though it may be possible to fit a lossy interpretable model to proxy the causal forest, in most cases, there is probably a much better alternative. At least in the case of this application, there is a stark trade-off between interpretability and performance.

\subsubsection{Using a best linear projection} \label{sec:blp}
One approach to fitting an interpretable model that draws on the power of the causal forest is to fit a best linear predictor (BLP). The BLP is a relatively well discussed approach compared to others discussed in this paper. It was an approach proposed by \citet{Semenova2021Debiased}, incorporated into the grf package and applied by \citet{Athey2019Estimating} in the most widely cited application of the causal forest. Undoubtedly the BLP adds value, particularly for hypothesis testing of results. However, its utility is situational. It assumes that a linear projection of the CATE onto a set of covariates (which may or may not be the predictors used in the causal forest) is useful in some way. This may mean that it meets the  assumptions neccessary for linear regression with valid standard errors, however it may be useful in a less formal way, for example in indicating potential drivers of heterogeneity which can then be explored in other ways. It also has the potential however to be misleading if used as the only way to try and understand treatment effect variation (it is easy to imagine for example very heterogeneous nonlinear distributions of CATEs that might produce a linear model with slope coefficients close to zero). It goes without saying that projecting effects with high-dimensional interactions between variables onto a combination of these variables will miss these important interaction effects.

One possible alternative is more flexible methods for trying to model effects. One approach to this is to use LASSO for selection of variables here in cases with a reasonably large number of covariates \citep{bahamyirou_doubly_2022}. It is worth noting that such an approach has no need for a final-stage causal forest and can simply be fit on the same doubly-robust scores used in estimation. Another possibility is the clustering approach used in the \textit{MCF} package \citep{lechner_modified_2019}.

Because its utility depends on the situation, BLP should work well as a tool for usability of the causal forest when used responsibly (i.e. when assumptions hold). The BLP was invented for this kind of application and allows researchers (or policy analysts) more familiar with linear models to make sense of the complex causal forest. However, they offer less value as accountability tools. The reason for this is that taken alone they provide little insight into the workings of a model that may lead to unjust outcomes (unless policy was made solely on the basis of the BLP).

Table \ref{tab:blp} shows the regression output for the BLP of returns on education in Australia, that a model that tries to predict doubly robust scores from covariates --- in this case the five most important predictors in the causal forest (these are selected per the procedure suggested in \citet{Athey2019Estimating} of selecting above mean importance variables to prune features, though this context is different from the one they discuss). Categorical variables were coded as dummies. Categories with less than 100 answers were excluded again to keep the regression output relatively small.

We see in Table \ref{tab:blp} that there are four variables with statistically significant results. Father's occupation on the AUSE106 scale which codes the status of all occupations in Australia from labourers to medical doctors. Here returns on education grow with the status of the participant's father's occupation. Those born later saw lower returns on education, either because higher levels of education are worth less in income now than they used to be or because these people have had less time to realise the value of their human capital. Females have lower returns on education, perhaps because of the overall wage gap and in particular the effect of motherhood on wages. Finally, being born in the Phillippines greatly boosts returns on education. These findings seem on their face credible and we have been able to extract simple, testable hypotheses from the causal forest.

It is also worth noting a BLP can be fit directly on doubly robust scores without a causal forest involved (although fitting a BLP through the \textit{grf} package is helpful as jackknifing errors with the structure of the ensemble is a computationally cheap way to estimate standard errors where otherwise nuisance models would need bootstrapping).

\begin{table}[!htbp] \centering 
  \caption{Results of BLP regression model} 
  \label{tab:blp} 
\begin{tabular}{@{\extracolsep{5pt}}lc} 
\\[-1.8ex]\hline 
\hline \\[-1.8ex] 
 & \multicolumn{1}{c}{\textit{Dependent variable:}} \\ 
\cline{2-2} 
\\[-1.8ex] &   \\ 
\hline \\[-1.8ex] 
 Father's occupation on AUSEI06 occupational status scale & 39.277$^{**}$ \\ 
  & (15.677) \\ 
  & \\ 
 Date of birth & $-$0.130$^{***}$ \\ 
  & (0.040) \\ 
  & \\ 
 Sex --- Female & $-$2,954.175$^{***}$ \\ 
  & (712.840) \\ 
  & \\ 
 Country of birth --- Australia & 867.550 \\ 
  & (1,487.134) \\ 
  & \\ 
 Country of birth --- China & 3,545.617 \\ 
  & (5,243.619) \\ 
  & \\ 
 Country of birth --- India & 2,000.223 \\ 
  & (3,628.917) \\ 
  & \\ 
 Country of birth --- New Zealand & $-$964.485 \\ 
  & (2,538.236) \\ 
  & \\ 
 Country of birth --- Philippines & 7,254.726$^{***}$ \\ 
  & (2,295.799) \\ 
  & \\ 
 Country of birth --- United Kingdom & 3,966.406$^{*}$ \\ 
  & (2,047.823) \\ 
  & \\ 
 Country of last year of school --- Australia & 2,815.494 \\ 
  & (1,901.103) \\ 
  & \\ 
 Country of last year of school --- India & 3,791.466 \\ 
  & (7,212.391) \\ 
  & \\ 
 Country of last year of school --- New Zealand & 5,563.705 \\ 
  & (3,420.876) \\ 
  & \\ 
 Country of last year of school --- United Kingdom & $-$2,229.539 \\ 
  & (2,348.041) \\ 
  & \\ 
 Constant & 2,430.163 \\ 
  & (1,710.112) \\ 
  & \\ 
\hline \\[-1.8ex] 
\hline 
\hline \\[-1.8ex] 
\textit{Note:}  & \multicolumn{1}{r}{$^{*}$p$<$0.1; $^{**}$p$<$0.05; $^{***}$p$<$0.01} \\ 
\end{tabular} 
\end{table} 
\label{table:blp}

\FloatBarrier
\subsection{Identifying problems in nuisance models}
While this section has for the most part focused on transparency in the heterogeneity model, it is worth discussing briefly transparency in the nuisance models. The transparency needs in nuisance models are different from those in heterogeneity models. In particular, there is no usability need to actually understand how nuisance models are making their predictions, if the models are removing confounding correctly there is no usability concern in understanding them. If they are not, it is not particularly important we understand how the model work, but simply that we either improve model fit or take a different approach to identification. In accountability concerns, the important thing is that we correctly identify effects, there is no equity concern outside of models performing identification poorly in ways that might harm people for whom identification is poor \citep{rehill_fairness_2023}. However, this is not a case of fair outcomes being counter to the goal of modellers, rather good identification is something that all parties want and which can be achieved through a simpler set of tools. Here we look at refutation tests as one way to do this for a continuous treatment.

\subsubsection{Refutation tests}
Refutation tests could be considered a kind of narrow causal AI algorithmic transparency tool. These are diagnostic tools developed for generic causal modelling approaches \citep{Sharma2021DoWhy:}. While they do not give much insight into the underlying causal effects, they are very useful to diagnosing problems in modelling either for users who want to make sure their models meet their assumptions or for those who wish to critique models. They are tests of the underlying causal assumptions of a model, however, they can also diagnose problems with the nuisance modelling approaches insofar as nuisance models may be failing to properly model out confounding effects. In this way they refute analysis but do not neccessarily provide a good idea of how one might fix the problem of confounding.

We test two particular refutation approaches by randomly shuffling treatment and then outcomes. This functions as a version of the placebo treatment and dummy outcome test that preserves the underlying univariate distribution while rendering it independent from other variables \citep{Sharma2021DoWhy:}. These estimates are plotted against the real treatment or outcome models to check if poor nuisance model fit is affecting treatment effect estimates. We then look at whether treatment effect estimates move to zero. Random treatment should in a correctly specified model drive the ATE and CATE estimates to zero \citep{Sharma2021DoWhy:}. Table \ref{table1} records the average treatment effect models for these tests. The outcome nuisance model (tested via randomised treatment) seems to be performing well, the treatment model (tested via randomised outcome) less so (although we would still fail to reject the null that the point estimate differs from zero at $p=0.05$).

\renewcommand{\arraystretch}{1.3}
\begin{table}[]
\caption{\label{table1}Results of refutation tests}
\begin{center}
\begin{tabular}{|l|l|}
\hline
            & ATE estimate  \\ \hline
Randomise w & \$9 (\$256)     \\ \hline
Randomise y & \$-716 (\$535)  \\ \hline
\end{tabular}
\end{center}
\end{table}

This can also be used to examine HTE estimates. When looking at the effect across different treatment and outcome levels with the random treatment, we still see a large amount of noise but also, the effect across different levels of education is on average near zero, suggesting orthogonalisation to outcome is working well. This is not the case for orthogonalisation to  which means that the outcome model is performing well but the treatment model is not.

This test helps us understand how the model is working and diagnose an identification problem that is not clear when just reviewing ATE estimates. Of course, in theory a researcher could simply find this particular problem when reviewing nuisance function fit in the first place, but the refutation test gives a clearer picture of where lack of fit is leading to bias in the causal estimate and where it has no effect (after all, there are two nuisance models with interacting effects on the final estimate). In addition, the autoML features in the grf library and the fact nuisance functions are fit automatically by default mean that in practice there is likely to not be much attention paid to the fit of nuisance function in causal forest modelling compared to traditional machine learning applications.

\begin{figure}
\begin{tabular}{cc}
\subfloat[Treatment estimates for placebo treatment by actual treatment level]{\includegraphics[width = 3in]{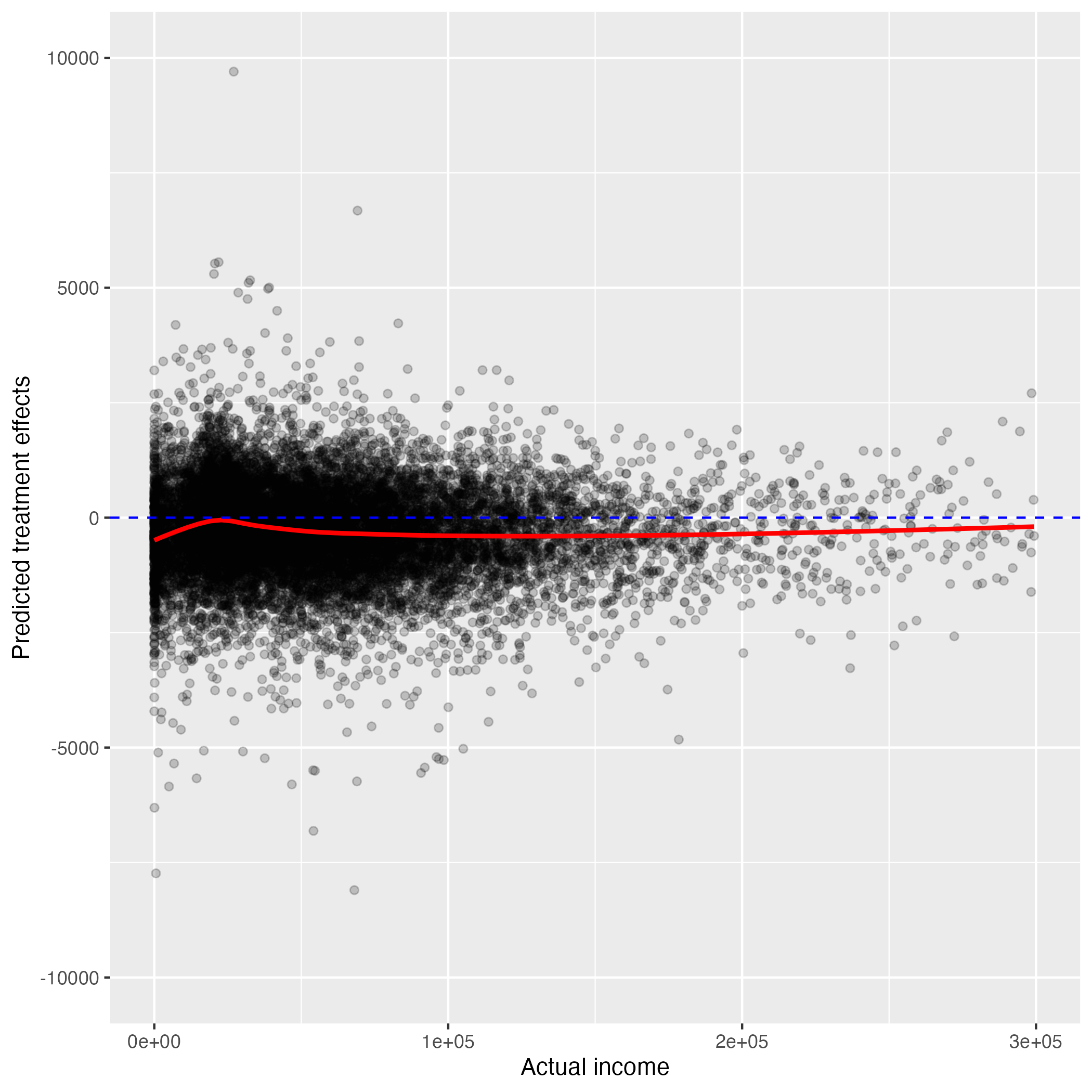}} &
\subfloat[Treatment estimates for placebo treatment by actual outcome]{\includegraphics[width = 3in]{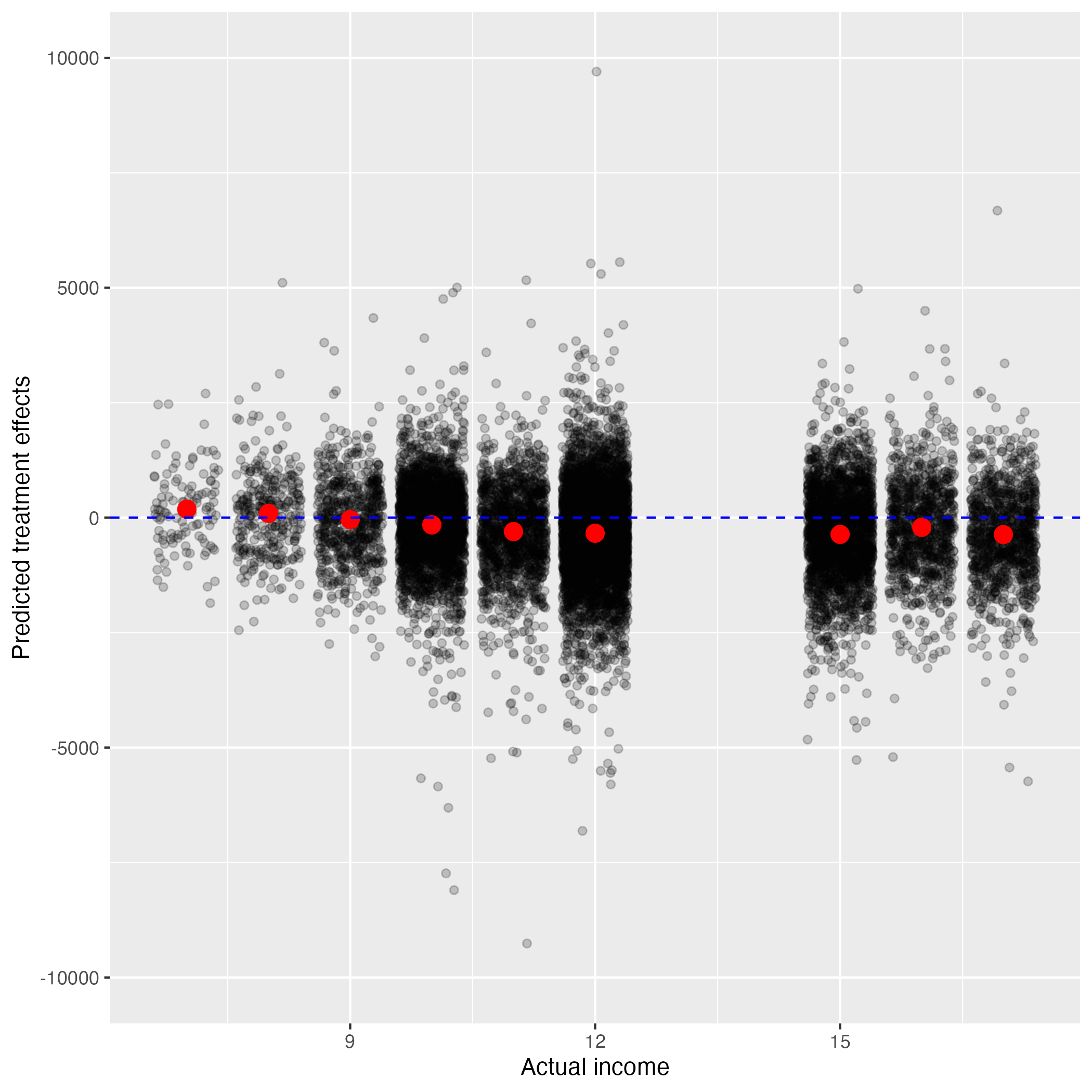}} \\
\subfloat[Treatment estimates for randomised outcome by actual treatment]{\includegraphics[width = 3in]{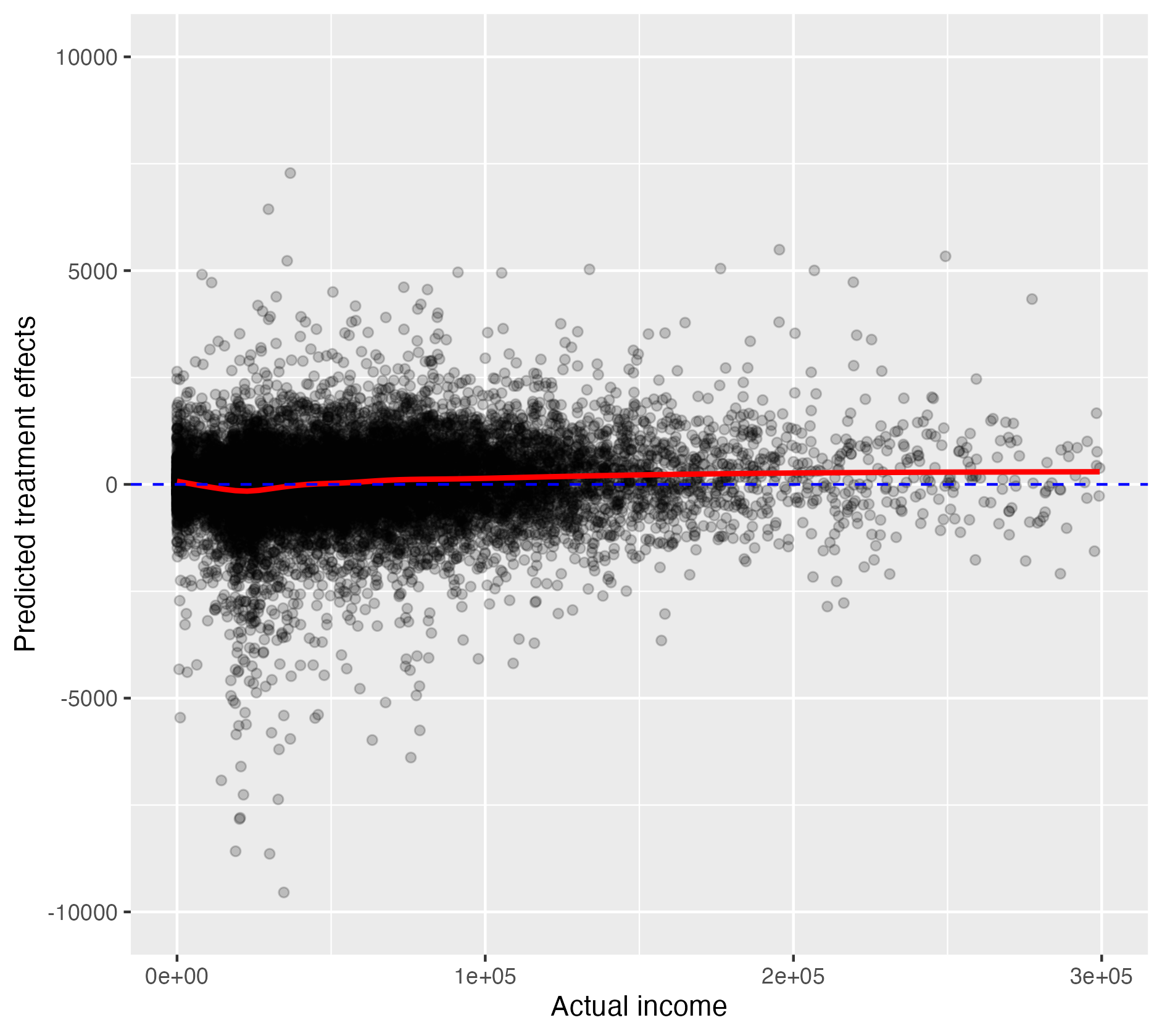}} &
\subfloat[Treatment estimates for randomised outcome by actual outcome]{\includegraphics[width = 3in]{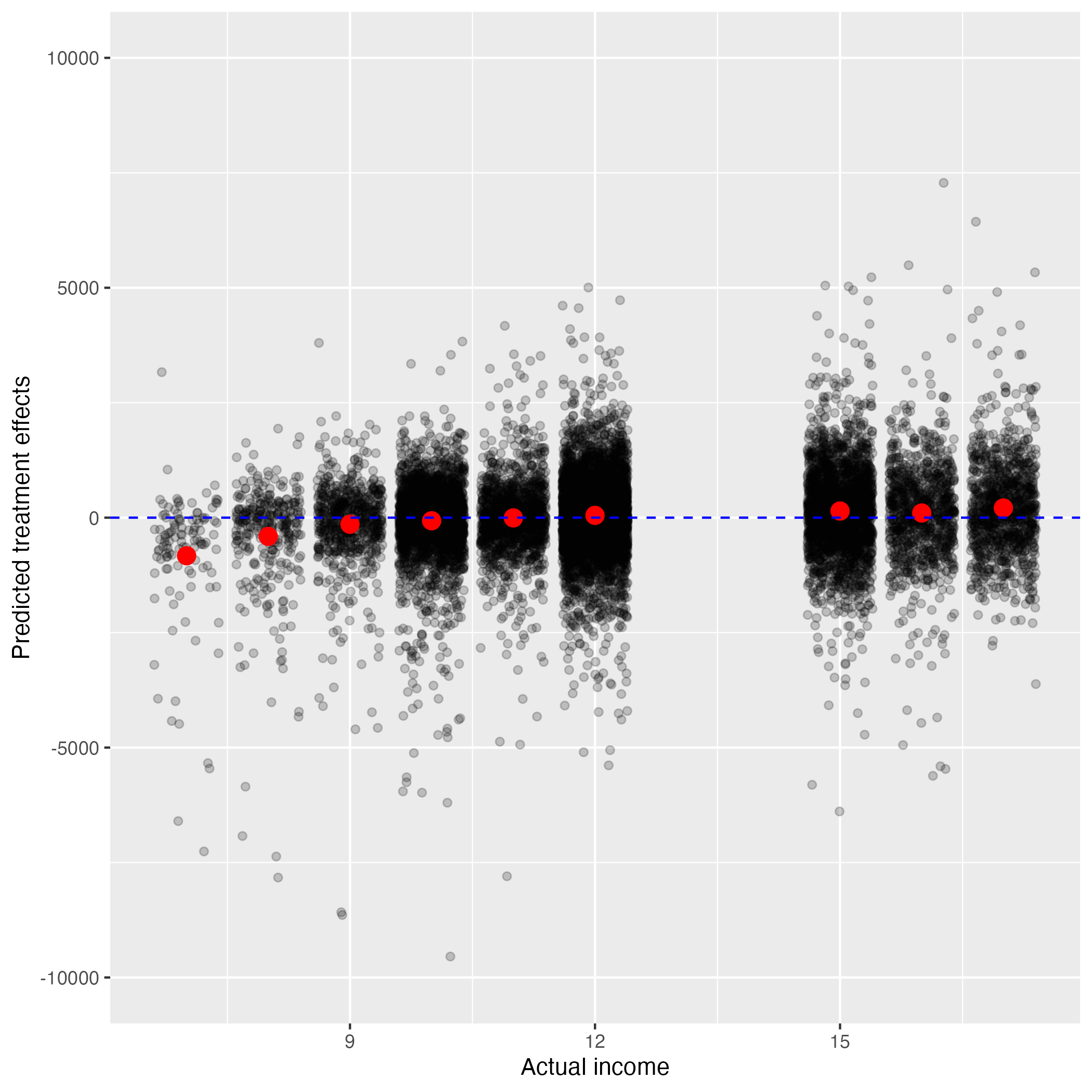}}
\end{tabular}
\caption{Effect of refutation tests on estimated treatment effects (treatment effects should be near zero, conditional averages are averages of doubly robust scores, not the individual estimates shown as points).}
\label{fig:placebo}
\end{figure}

Figure \ref{fig:placebo} shows the refutation tests seem to work fairly well across the distribution of outcome and treatment. It seems like the causal effect is well identified.

This kind of analysis is mostly useful for usability. The reason for this is that running placebo tests would ideally be part of diagnosing a model's problems before it is used \citep{Sharma2021DoWhy:}. While identification problems could pose justice issues that might be of interest for accountability, the problem is that confounding is not really a problem of the individual or small-group kind accountability involves. To the extent there are confounding issues not identified by analysts at the time the model is fit, it is hard to understand whether the causal effect is estimated better or worse for certain individuals across all the possible covariates. It is equally hard to comment on the extent to which such confounding effected an unjust outcome.

\FloatBarrier

\section{Conclusion}
Causal machine learning offers tremendous promise to researchers tackling specific kinds of research questions, but transparency poses a real problem both to users of the model and those who might want to hold these users accountable. This paper has laid out some issues around the use of causal machine learning in policy research around government decisions, but it is far from a full survey of all the possible issues that might arise. A much larger body of knowledge is needed to properly establish best practice for the use of these methods. In particular studies on the use of causal machine learning methods applied to real-world policy problems as these case studies become available would be useful, as would experiments on how causal machine learning analysis affects decision-making when compared to traditional quantitative methods along the lines of \citet{Green2019Disparate} or \citet{Logg2019Algorithm} on predictive systems.

In the absence of an significant existing critical literature, we offer the following two guidelines for using causal machine learning. Firstly, causal machine learning should only be used where it is additive to the evaluation. By that we mean there is a reason to use a powerful but black-box method over a less powerful but interpretable method (i.e. standard policy evaluation methods). The obvious reason to do this in the case of the causal forest is when it will be valuable to understand heterogeneous effects at the individual level or where we have little theoretical knowledge about the drivers of heterogeneity and so have to undertake a data-driven exploration of these effects instead \citep{Athey2018Machine}. Causal machine learning methods should not be used for high-stakes policy evaluation simply because they are novel or because it is inconvenient to find quasi-experimental / experimental data. The decision to use them should be made understanding that there will likely be responsibility gaps in the use of such novel methods in policy-making \citep{Matthias2004responsibility,Olsen2017Democratic}. Secondly, causal machine learning should meet the standards of transparency expected from predictive machine learning or traditional causal modelling in government. Just because the nature of the analysis is different does not mean the same issues that currently plague machine learning models used in government are not a concern for causal models.

Within this paper, we have tried to establish why transparency is important in causal modelling, analogising this to predictive applications. However, we have also laid out how these two types of analysis bring with them different transparency needs. Causal machine learning lays out the data-generating process of the underlying data and the fits into a process of human decision-making with well-established transparency requirements. On the other hand, these models usually involve fitting several black-box models where even XAI and IAI approaches fail to explain much about how all these models relate to each other (there are not even good procedures for estimating error through the process). More work is needed to tailor tooling to these different transparency implications.

\section*{Acknowledgements}
Patrick Rehill is supported by an Australian Government Research Training Program Scholarship.

\section*{Data availability statement}
The data used for the case study in this paper are available from the Australian Data Archive at doi:10.26193/KXNEBO. They require an application to access. The code used is available at \url{https://github.com/pbrehill/CML-transparency-justice}.

\section*{Declaration of Competing Interest}
Patrick Rehill has no competing interests in relation to this work.

Nicholas Biddle has no competing interests in relation to this work.

\section*{Funding statement}
This work received no specific grant from any funding agency, commercial or not-for-profit sectors.

%Bibliography
\bibliographystyle{agsm}
\setcitestyle{authoryear,open={(},close={)}}
\bibliography{references}

\section*{Appendix A --- Variables used in application}

\begin{tabular}{lr}
\toprule
Name & Variable importance\\
\midrule
HF5 Sex & 0.0166132\\
HF6 Date of Birth & 0.0544183\\
History: Country of birth & 0.0098066\\
History: Country of birth - brief & 0.0000000\\
\addlinespace
History: Aboriginal or Torres Strait Islander origin & 0.0041727\\
History: Were you living with both your own mother and \\
\quad\quad\quad father around the time you were 14 years old & 0.0031583\\
History: Did your mother and father ever get divorced or
separate & 0.0310775\\
History: How old were you when you first moved out of home
as a young person & 0.0004202\\
History: Ever had any siblings & 0.0146042\\
\addlinespace
History: How many siblings & 0.0050177\\
History: Were you the oldest child & 0.0728535\\
History: Father's Country of Birth & 0.0150673\\
History: Mother's Country of Birth & 0.0131440\\
History: How much schooling father completed & 0.0145330\\
\addlinespace
History: How much schooling mother completed & 0.0026701\\
History: Father completed an educational qualification after
leaving school & 0.0025757\\
History: Mother completed an educational qualification after
leaving school & 0.0001508\\
History: Was father in paid employment when you were 14 & 0.0228523\\
History: Father's occupation - 4 digit ANZSCO 2006 & 0.0060168\\
\addlinespace
History: Father's occupation 2-digit ANZSCO 2006 & 0.0013925\\
History: Was father unemployed for 6 months or more while
you were growing up & 0.0051541\\
History: Was mother in paid employment when you were 14 & 0.0596648\\
History: Mother's occupation - 4 digit ANZSCO 2006 & 0.0248035\\
History: Mother's occupation 2-digit ANZSCO 2006 & 0.0104738\\
\addlinespace
History: Country of last school year & 0.0126794\\
History: Age retired/intends to retire & 0.0842183\\
History: Time since FT education - years & 0.1488299\\
History: AUSEI06 occupational status scale, Father's
occupation & 0.0555382\\
History: AUSEI06 occupational status scale, Mother's
occupation & 0.0314822\\
\addlinespace
History: ISCO-88 from ANZSCO 2006, Father's occupation & 0.1045225\\
History: ISCO-88 from ANZSCO 2006, Mother's occupation & 0.0082781\\
History: ISCO-88 from ANZSCO 2006 2-digit, Father's
occupation & 0.0227356\\
History: ISCO-88 from ANZSCO 2006 2-digit, Mother's
occupation & 0.0012949\\
History: Father's occupation 1-digit ANZSCO 2006 & 0.0034315\\
\addlinespace
History: Mother's occupation 1-digit ANZSCO 2006 & 0.1363484\\
\bottomrule
\end{tabular}

\section*{Appendix B --- Additional waterfall plots of random respondent SHAP values}

\begin{figure}[!h]
    \caption{Waterfall plot of SHAP values for random respondent 2}
    \centering
    \includegraphics[scale=0.40]{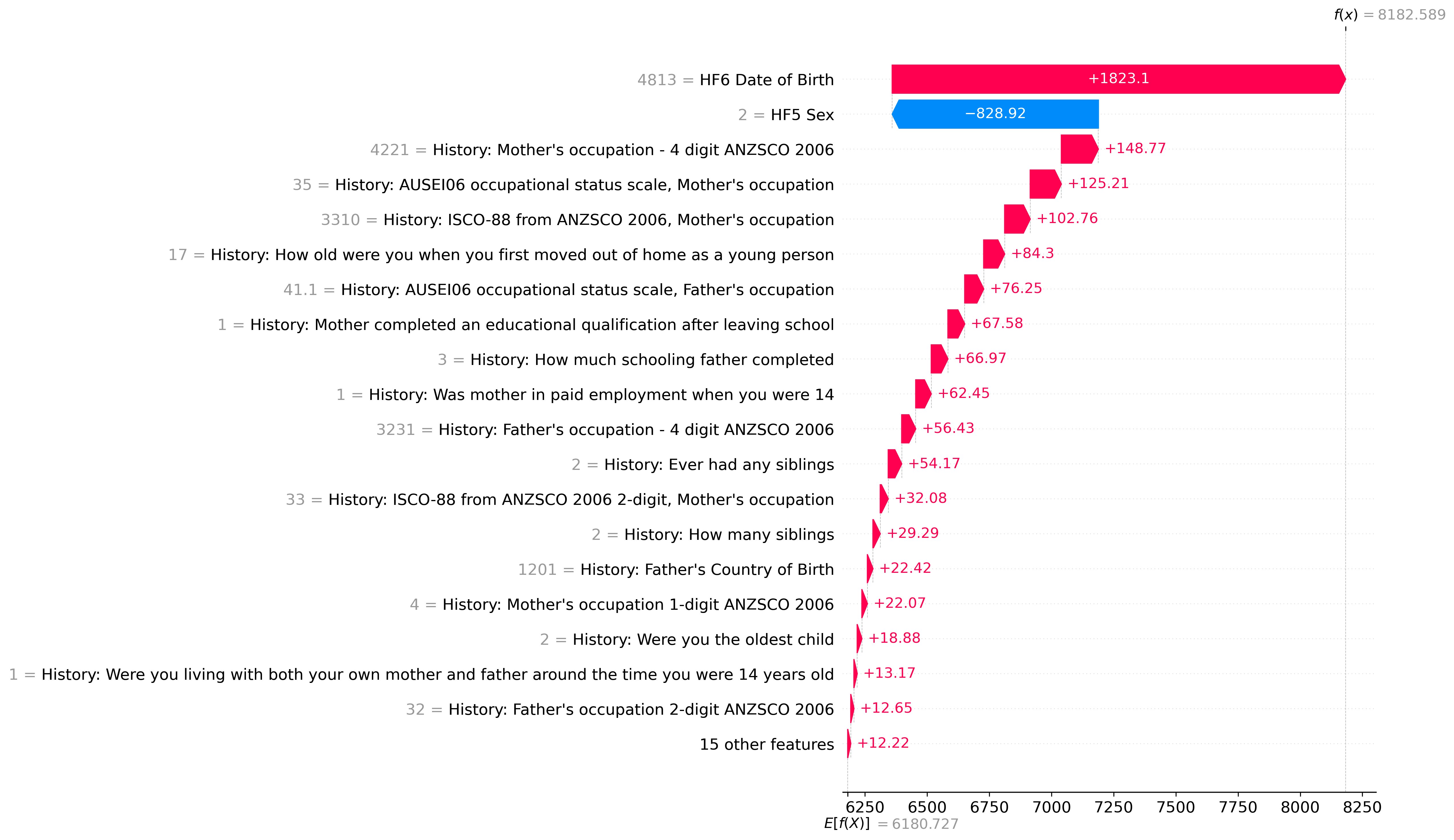}
    \label{fig:a1}
\end{figure}

\begin{figure}[!h]
    \caption{Waterfall plot of SHAP values for random respondent 3}
    \centering
    \includegraphics[scale=0.40]{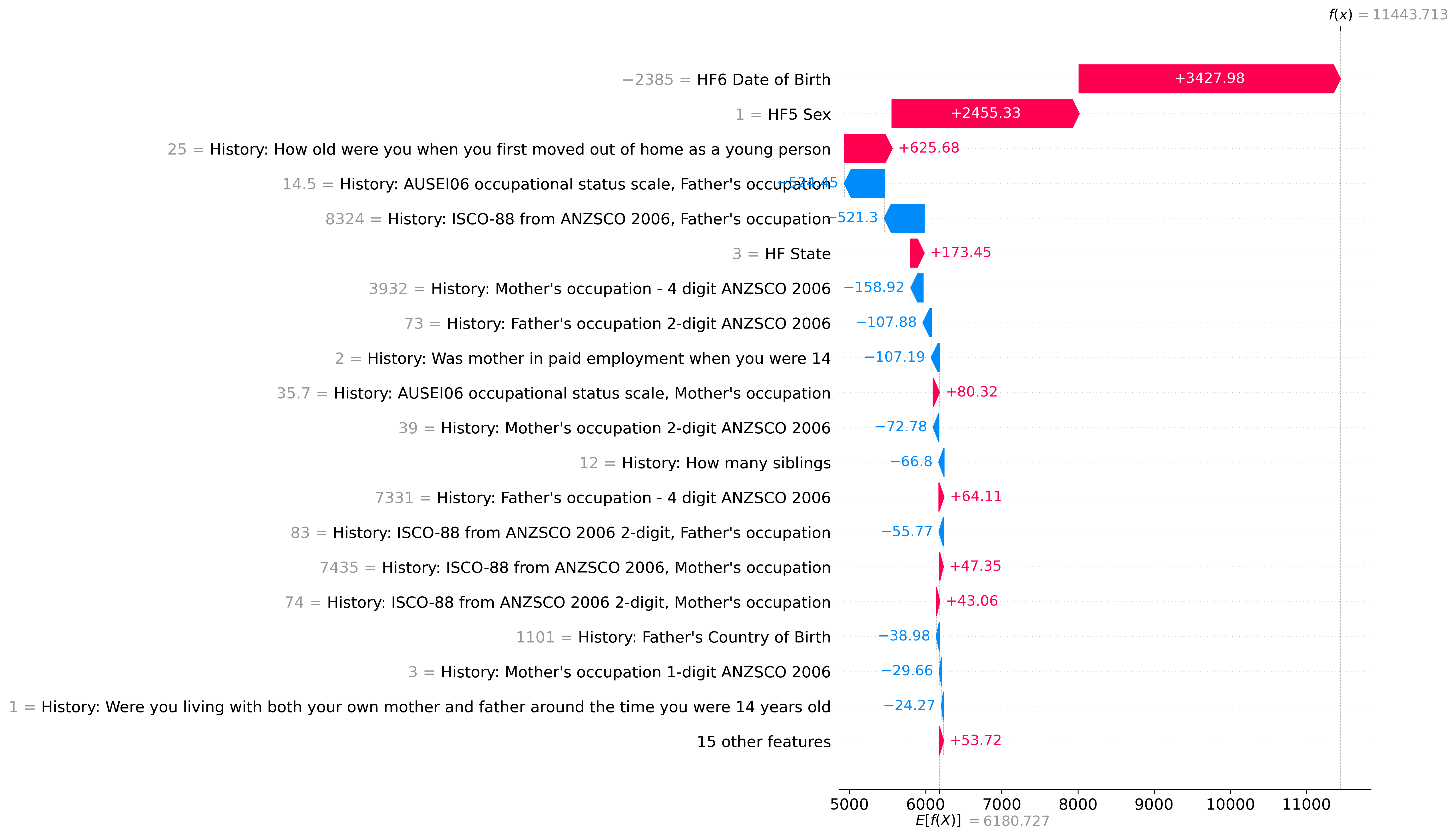}
    \label{fig:a2}
\end{figure}

\begin{figure}[!h]
    \caption{Waterfall plot of SHAP values for random respondent 4}
    \centering
    \includegraphics[scale=0.4]{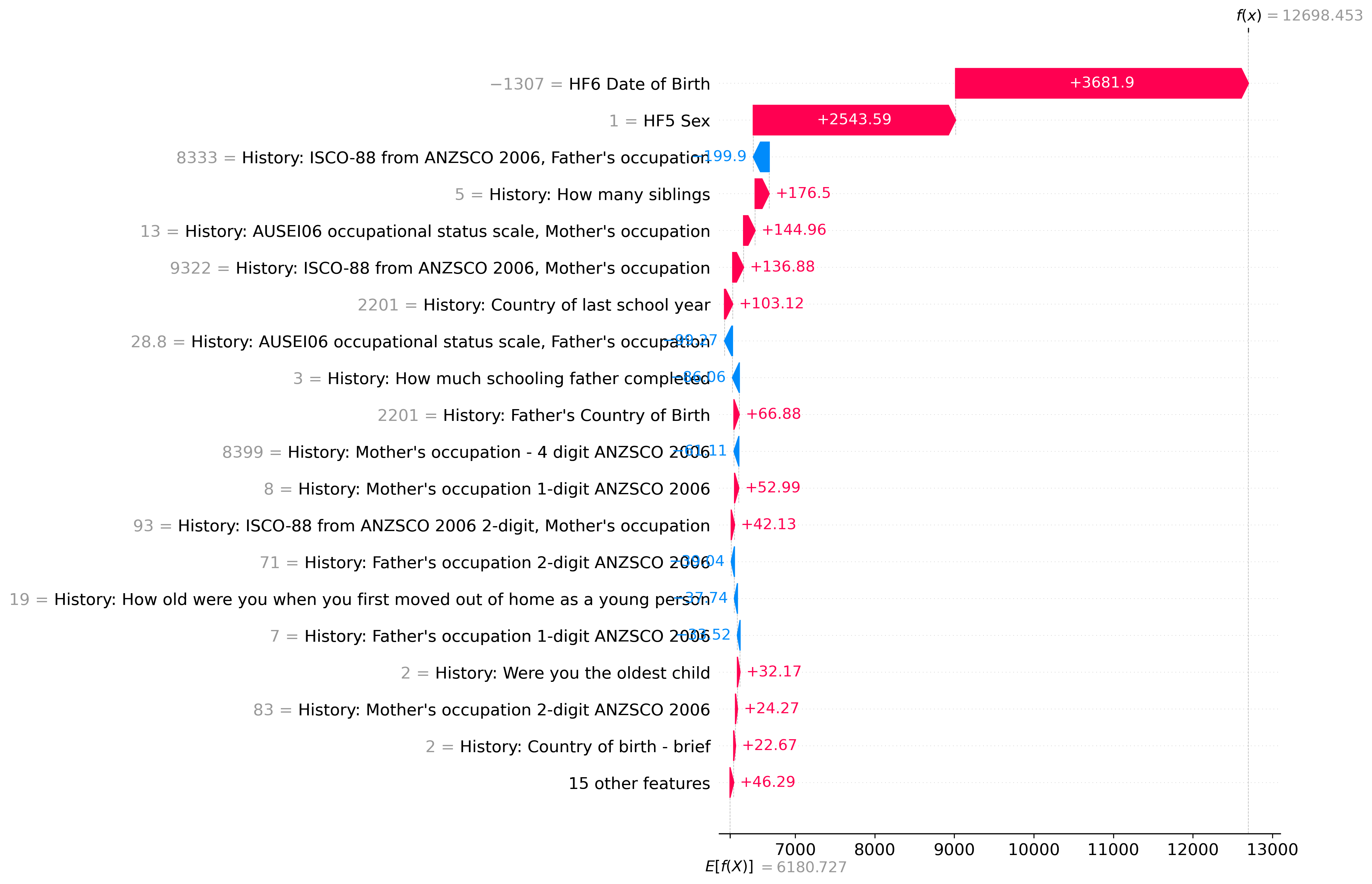}
    \label{fig:a3}
\end{figure}

\begin{figure}[!h]
    \caption{Waterfall plot of SHAP values for random respondent 5}
    \centering
    \includegraphics[scale=0.4]{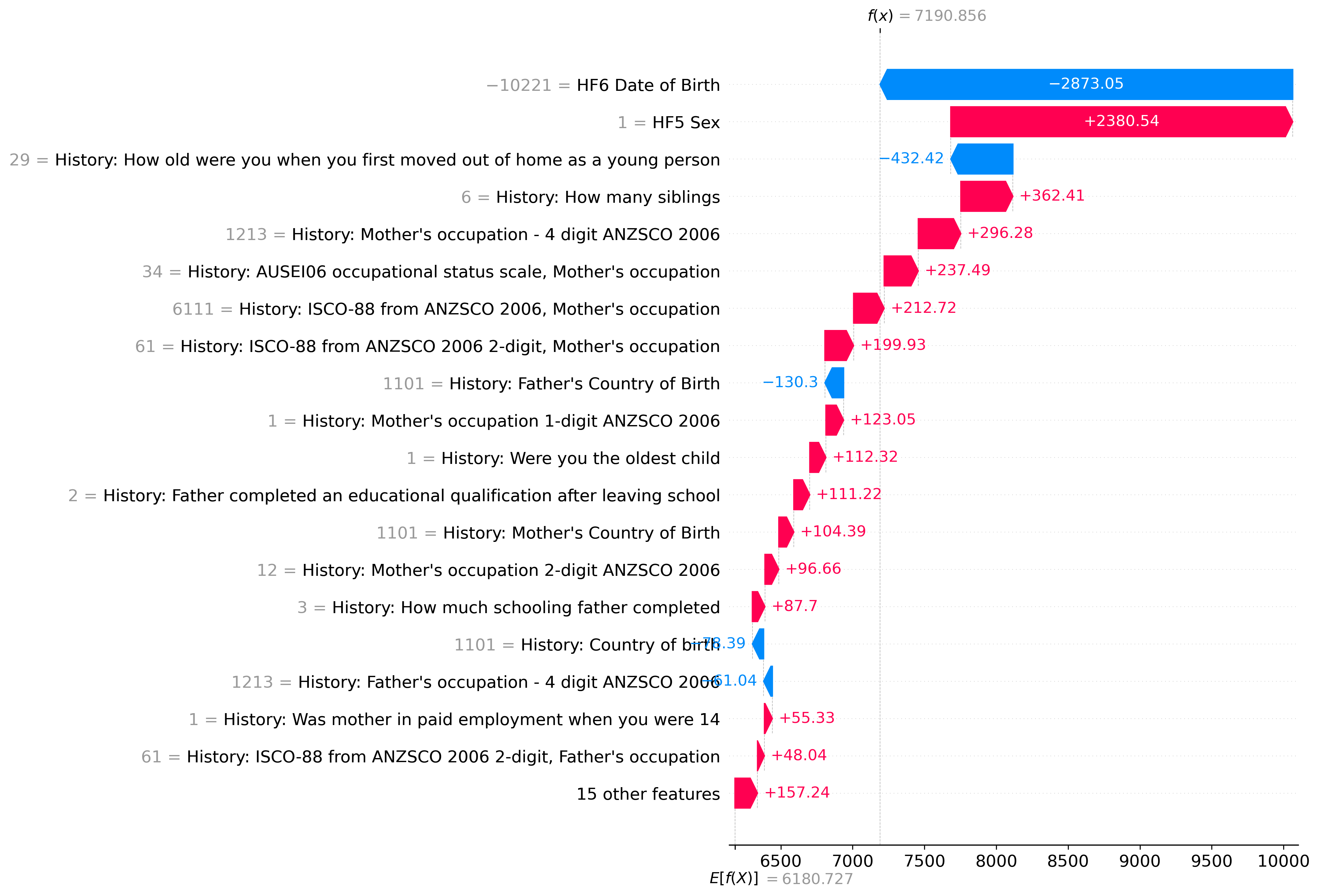}
    \label{fig:a4}
\end{figure}

\section*{Appendix C --- Table of definitions} \label{sec:appendixc}
% \usepackage[normalem]{ulem}
\begin{longtable}{>{\raggedright}p{4cm} >{\raggedright\arraybackslash}p{8cm} >{\raggedright\arraybackslash}p{3cm}} 

\toprule
\textbf{Term} & \textbf{Explanation} & \textbf{Further Reading} \\
\midrule
\endfirsthead

\toprule
\textbf{Term} & \textbf{Explanation} & \textbf{Further Reading} \\
\midrule
\endhead

Best Linear Projection (BLP)              & A best linear projection projects doubly robust scores onto a linear model. This is helpful because of treatment effect heterogeneity is linear this can allow one to identify important predictors and also to hypothesis test linear relationships. & \citet{Semenova2021Debiased}\\
Black-box Models                          & Models where the internal logic or decision-making process is not easily understandable.                                                                                                                                                                                                                                                                                                                                                                                                                                                                                                                                                                                                                                                                                                                                                                                                                                                                                                                                                                                                                                                                                                                                                                                                               & \citet{Lipton2018Mythos}                 \\
Causal Forest                             & A popular method for estimating heterogeneous treatment effects in causal machine learning. To simplify how the forest works, it takes removes selection effects by trying to predict treatment and outcome using nuisance models (like Double Machine Learning). It then takes the residuals of these models (which removes selection effects) and plugs them into a random forest model made up of causal trees. These trees are designed to split to maximise within-node treatment effect heterogeneity and they use 'honest splitting' prevent over-fitting and give asymptotically normal predictions. This means half the data is used to split the tree and the other half is used to estimate effects in leaf nodes. This ensemble is then used not to directly predict but, to create an 'adaptive kernel' which is then used with a doubly robust estimator (augmented inverse probability weighting by default) to get an estimate. The doubly robust estimator uses doubly robust scores estimated from the nuisance model and takes an average of these weighted by kernel distance from each training example. This means that for a new datapoint with X=x, we weight based on the fraction of the time that each training example would end up in the same leaf as the new datapoint. & \citet{Athey2019Generalized,Athey2019Estimating,Wager2018Estimation}                \\
Causal Machine Learning                   & Machine learning methods designed to estimate causal effects rather than simply predict outcomes. Unlike predictive models, they are fitting a fundementally unknowable quantity (because the treatment effect is the difference between two potential outcomes that cannot both be observed).                                                                                                                                                                                                                                                                                                                                                                                                                                                                                                                                                                                                                                                                                                                                                                                                                                                                                                                                                                                                         &   \citet{Lechner2023Causal}              \\ 
Double Machine Learning                   & A method for removing selection effects from data in causal inference using predictive machine learning models as nuisance models. Essentially it involves using these two nuisance models, one to predict treatment, one to predict outcome then taking the residuals from these models and feeding them into an estimator of some sort. So long as models are cross-fit (or more simply, predictions are made out-of-sample) using machine learning models will not have a biasing effect. Theoretically, this process of taking residuals removes selection effects from the data (if there are no unobserved confounders).                                                                                                                                                                                                                                                                                                                                                                                                                                                                                                                                                                                                                                                                                                                                                                                                                                                                                         &   \citet{Chernozhukov2018Double/debiased}              \\
Explainable AI (XAI)                      & Techniques to make machine learning models understandable to humans that involve explaining predictions made by a black box. This often means fitting a secondary model on the predictions of the black-box. These explanations are often 'local' in the sense that we cannot understand how the model would make predictions for any data point from an explanation.                                                                                                                                                                                                                                                                                                                                                                                                                                                                                                                                                                                                                                                                                                                                                                                                                                                                                                                                  &  \citet{Lipton2018Mythos}               \\
Heterogeneous Treatment Effect Estimation & Estimating how a treatment affects different units differently. In practice we are generally estimating a conditional average treatment effect (CATE) i.e. an estimate of treatment effect given certain characteristics X but other terms like group average treatment effect (GATE) or individual treatment effect (ITE) are sometimes used as well.                                                                                                                                                                                                                                                                                                                                                                                                                                                                                                                                                                                                                                                                                                                                                                                                                                                                                                                                                 &  \citet{Athey2019Estimating,Nie2020Quasi-Oracle,Künzel2019Metalearners}               \\
Interpretable AI (IAI)                    & Techniques involving fitting 'white-box' models which may be simpler than 'black-box' models, but allow a human to get a global understanding of the model. This means a human can often perform inference themselves seeing an interpretable model (for example tracing a path from root to node on a decision tree). Some interpretable AI approaches simply fit an interpretable model first while others 'distill' interpretable models from more complex black-box models in some way.                                                                                                                                                                                                                                                                                                                                                                                                                                                                                                                                                                                                                                                                                                                                                                                                            &  \citet{Rudin2019Stop}               \\
Nuisance Models                           & Models that do not predict the quantity that is of interest in a study but rather are a necessary intermediate step in getting to modelling that final quantity. Here we have nuisance models for the purposes of identifying a causal effect (see Double Machine Learning).                                                                                                                                                                                                                                                                                                                                                                                                                                                                                                                                                                                                                                                                                                                                                                                                                                                                                                                                                                                                                           & \citet{Chernozhukov2018Double/debiased}                \\
Predictive Machine Learning               & Machine learning methods designed to predict an outcome from data.                                                                                                                                                                                                                                                                                                                                                                                                                                                                                                                                                                                                                                                                                                                                                                                                                                                                                                                                                                                                                                                                                                                                                                                                                                     &  \citet{Athey2017State}               \\
R-Loss Function                           & A loss function for heterogeneous treatment effect estimation (through the R-Loss meta-learner). This is most commonly used as the objective function in the heterogeneity model of the causal forest.                                                                                                                                                                                                                                                                                                                                                                                                                                                                                                                                                                                                                                                                                                                                                                                                                                                                                                                                                                                                                                                                                                 &            \citet{Nie2020Quasi-Oracle}     \\
SHAP (SHapley Additive exPlanations)      & An XAI method based on game theory for explaining a prediction by showing the impact that the values of some explanatory X variables had on the prediction.                                                                                                                                                                                                                                                                                                                                                                                                                                                                                                                                                                                                                                                                                                                                                                                                                                                                                                                                                                                                                                                                                                                                            &  \citet{lundberg_unified_2017}               \\
Variable Importance                       & A metric that shows how important different variables are in a model. In predictive modelling it is common to express this in terms of the amount each variable helps to minimise the loss function. In the causal forest, without a clear loss function (R-Loss is too noisy for a predictive-style approach to be practical) it is common to simply count the number of times the forest split on each variable. More advanced approaches have recently been proposed for the causal forest.                                                                                                                                                                                                                                                                                                                                                                                                                                                                                                                                                                                                                                                                                                                                                                                                         &  \citet{benard_variable_2023,Athey2019Estimating}              
\end{longtable}

\end{document}